\documentclass[letterpaper, 10 pt, conference]{ieeeconf}  

\IEEEoverridecommandlockouts                              

\usepackage{graphicx} 
\usepackage{amsmath} 
\usepackage{amssymb}  

\title{\LARGE \bf
Transform Invariant Auto-encoder
}

\author{Tadashi Matsuo$^{1}$, Hiroya Fukuhara$^{2}$ and Nobutaka Shimada$^{3}$
\thanks{*This work was supported by JSPS KAKENHI Grant Number 24500224, 15H02764 and MEXT-Supported Program for the Strategic Research Foundation at Private Universities, S1311039, 2013-2016.}
\thanks{$^{1,3}$Tadashi Matsuo and Nobutaka Shimada are with College of Information Science and Engineering, Ritsumeikan University, 1-1-1 Noji-higashi, Kusatsu, Shiga, JAPAN
        $^{1}${\tt\small matsuo@i.ci.ritsumei.ac.jp}}%
\thanks{$^{2}$Hiroya Fukuhara is with Graduate School of Information Science and Engineering, Ritsumeikan University, 1-1-1 Noji-higashi, Kusatsu, Shiga, JAPAN}%
}

\newcommand{\argmin}{\mathop{\rm argmin}\limits}

\newcommand{\defineequal}{\overset{\mathrm{def}}{=}}

\newcommand{\norm}[1]{\left\| #1 \right\|}

\begin{document}

\maketitle
\thispagestyle{empty}
\pagestyle{empty}

\begin{abstract}
 The auto-encoder method is a type of dimensionality reduction method.
 A mapping from a vector to a descriptor that represents
 essential information can be automatically generated from a set of
 vectors without any supervising information.
 However, an image and its spatially shifted version are encoded into
 different descriptors by an existing ordinary auto-encoder because
 each descriptor includes a spatial subpattern and its position.
 To generate a descriptor representing a spatial subpattern
 in an image,
 we need to normalize its spatial position in the images prior to
 training an ordinary auto-encoder; however, such a normalization is
 generally
 difficult for images without obvious standard positions.
 We propose a transform invariant auto-encoder and an inference model of
 transform parameters.
 By the proposed method, we can separate an input into a transform
 invariant descriptor and transform parameters.
 The proposed method can be applied to various auto-encoders without
 requiring any special modules or labeled training samples.
 By applying it to shift transforms, we can achieve a shift invariant
 auto-encoder that can extract a typical spatial subpattern
 independent of its relative position in a
 window.
 In addition, we can achieve a model that can infer shift parameters
 required to restore the input from the typical subpattern.
 As an example of the proposed method,
 we demonstrate that a descriptor generated by a shift invariant
 auto-encoder
 can represent a typical spatial subpattern.
 In addition,
 we demonstrate the imitation of a human hand by a robot hand as an
 example of a regression based on spatial subpatterns.
\end{abstract}

 \section{INTRODUCTION}
 The auto-encoder
 method \cite{BALDI198953,Hinton504,DBLP:journals/corr/MakhzaniF13} is
 a type of dimensionality reduction method.
 It can extract essential information from a vector via general
 non-linear mapping.
 Moreover, a mapping from a vector to a descriptor representing
 essential information can be automatically generated from a set of
 vectors without any supervising information.
 %
 %

 %
 %
 In general, an auto-encoder is generated as an encoder and decoder
 pair.
 The encoder converts a vector to a descriptor with lower
 dimensionality,
 and the decoder approximately restores the original vector from the
 descriptor.
 An auto-encoder can be trained using a set of training samples by
 minimizing the restoration error of the encoder--decoder combination.
 After the training, the encoder should be able to generate a numerical
 representation of the principal components required to restore the
 original vector.
 Because the encoder and the decoder can be non-linear and can be
 trained without supervisor information,
 the auto-encoder method is suitable for allocating numerical vectors to
 targets without simple numerical representations.

\begin{figure}
  \vspace{.5em}
  {\centering
  \begin{minipage}[c]{.35\textwidth}
   {\centering
   \begin{tabular}{cc}
   \includegraphics[width=0.7\textwidth]{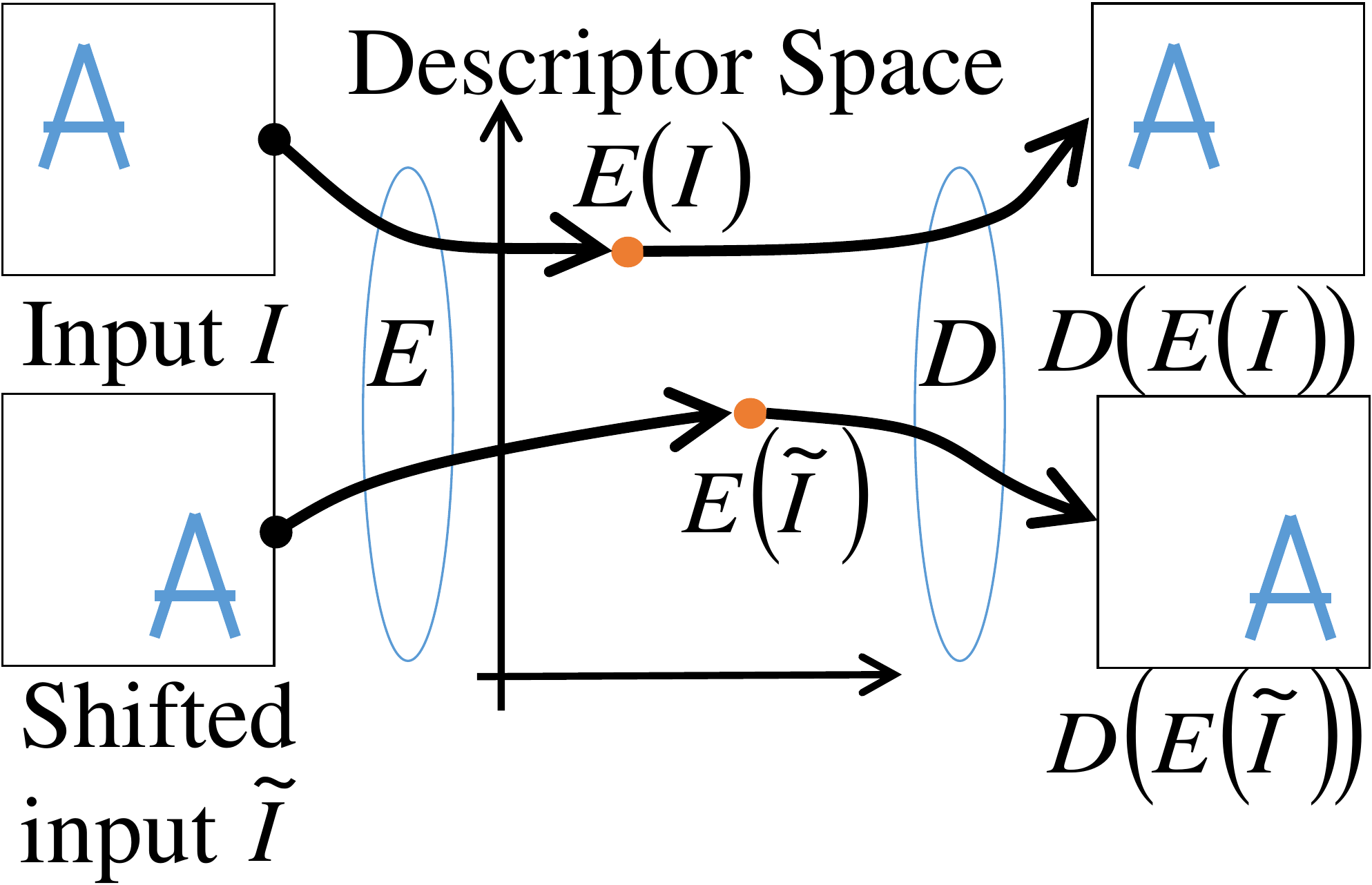}
   \\
   {\footnotesize (a) An ordinary auto-encoder}
   \\
   \includegraphics[width=0.7\textwidth]{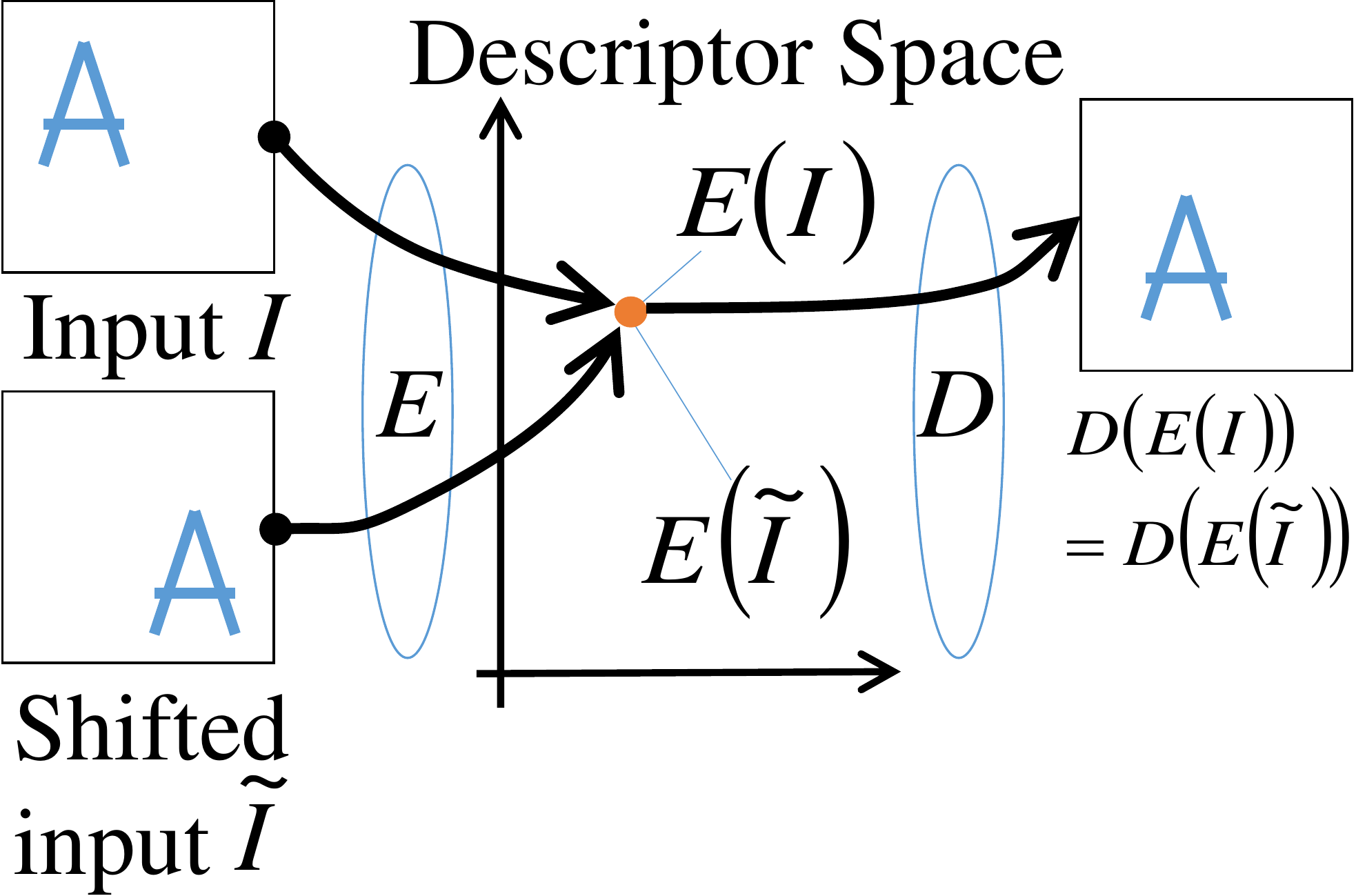}
   \\
   {\footnotesize (b) A shift invariant auto-encoder}
   \end{tabular}
   }
  \caption{Characteristics of auto-encoders}
  \label{fig:characteristics_of_auto_encoder}
  \end{minipage}
  \begin{minipage}[c]{.08\textwidth}
   \includegraphics[width=0.99\textwidth]{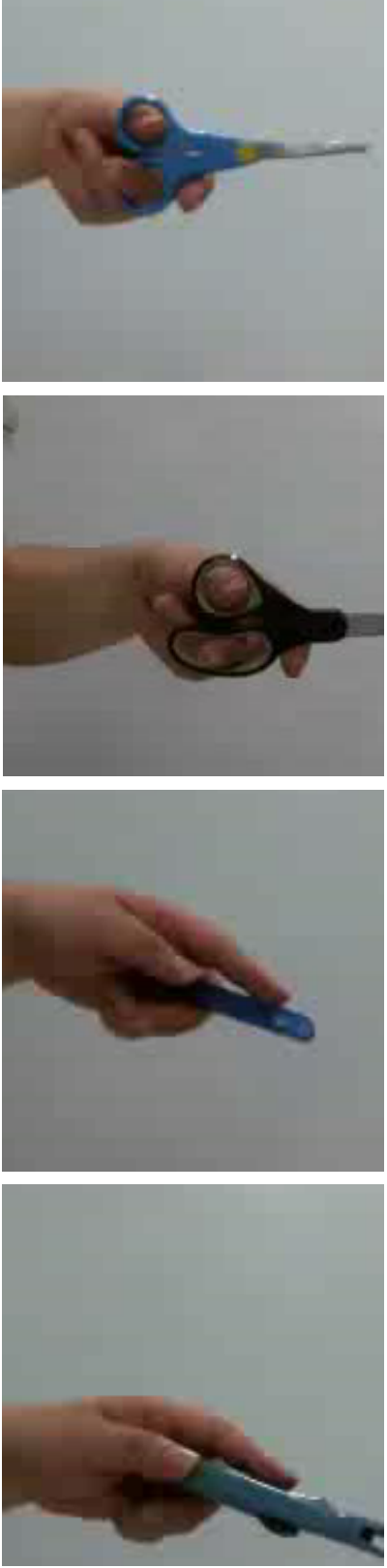}
   \caption{Appearances of hand--object interactions}
   \label{fig:appearances_of_hand_object_interactions}
  \end{minipage}
  }
 \end{figure}

 %
 %
 As a measure of the accuracy of the restoration performed by a
 auto-encoder, the simple
 $\ell^{2}$ measure is often used.
 Using the $\ell^{2}$ measure, an image and its spatially shifted
 version are considered to be different.
 If an auto-encoder is trained with the $\ell^{2}$ measure,
 images including a common spatial subpattern may be encoded as very
 different descriptors depending on the position of the subpatterns
 (Fig.~\ref{fig:characteristics_of_auto_encoder}(a)).
 This means that an ordinary auto-encoder inseparably embeds a spatial
 subpattern and its position within a descriptor.
 Therefore, to generate a descriptor representing a spatial subpattern
 in an image,
 we need to normalize its spatial position in the images prior to
 training an ordinary auto-encoder.
 However, such a spatial normalization is generally difficult.
 For example, the normalization of the appearances of various
 hand--object interactions such as those shown in
 Fig.~\ref{fig:appearances_of_hand_object_interactions},
 is not obvious and requires a pattern recognition technique to
 automatically find the standard for each image.

 %

 A combination of a convolutional neural network (CNN)\cite{726791} and
 spatial pooling ignores shifts of local small shifts, but it ignores
 only small shifts.
 M.~Jaderberg et al. proposed ``Spatial Transformer
 Networks'' \cite{DBLP:journals/corr/JaderbergSZK15}, which include a
 module to learn a spatial transform that is effective in
 classification.
 The transform module can cancel a transform
 including a spatial shift; however it must be trained with a teacher
 label
 for each input image.
 X.~Shen et al. proposed ``Transform-Invariant Convolutional Neural
 Networks'' \cite{Shen:2016:TCN:2964284.2964316}.
 but it requires a
 teacher label for each input image on the training process, too.
 M.~Baccouche et al. proposed
 ``Sparse Shift-Invariant Representation'' \cite{6460998},
 which requires a special training process where the best translation
 need to be found for each training sample.
 M.~Ranzato et al. proposed
 ``Sparse and Locally Shift Invariant Feature Extractor''\cite{4377108};
 however,
 its shift invariance is achieved at the cost of low spatial resolution
 by down sampling by max-pooling layer.

 We propose a transform invariant auto-encoder that outputs a descriptor
 invariant with respect to a set of transforms.
 By considering spatial shifts, the proposed method can generate a shift
 invariant auto-encoder, which extracts a typical spatial subpattern
 without regard to its relative position in a window
 (Fig.~\ref{fig:characteristics_of_auto_encoder}(b)).
 The proposed method is based on a novel cost function for training an
 auto-encoder, which
 induces transform invariance and accurate restoration.
 The proposed cost function is so designed as to be independent of
 the concrete structures of an encoder and a decoder of an auto-encoder.
 Therefore, it can be applied to various auto-encoders without requiring
 any special modules or labeled training samples.
 The proposed method can achieve transform invariance without requiring
 layers for low spatial resolution.
 Using the proposed method, we can encode spatial subpatterns in images
 even if the images are difficult to label or normalize, for example,
 the
 appearances of hand--object interactions.

 As an example, we have experimented with a shift invariant
 auto-encoder.
 In several experiments,
 we demonstrate that a descriptor using a shift invariant auto-encoder
 can represent a typical spatial subpattern.
 We also demonstrate the imitation of a human hand by a robot hand as an
 example of regression based on spatial subpatterns.

 \section{ORDINARY AUTO-ENCODER}
 In general,
 an auto-encoder is so trained that the encoder--decoder
 combination approximately restores an input in a certain input set.
 It is formulated as a problem minimizing a cost function
 $C_{\text{ord}}(E,D)$ defined as
 \begin{equation}
  \label{eq:1}
   C_{\text{ord}}\left(E, D\right)
   =
   \sum_{I \in S}\norm{I - D\left(E\left(I\right)\right)}^{2}_{2},
 \end{equation}
 where $S$, $D( \cdot )$, $E( \cdot )$, and $\norm{\cdot}_{p}$
 denote a set of inputs, the encoder, the
 decoder, and the $\ell^{p}$ norm, respectively.

 To minimize $C_{\text{ord}}(E,D)$,
 the decoder should be able to approximately restore an original vector
 $I$ from its descriptor $E(I)$,
 which has a lower dimensionality than $I$.
 By training the encoder $E$ and the decoder $D$ by minimizing
 $C_{\text{ord}}(E,D)$,
 information sufficient to restore an original vector can be extracted
 as a
 descriptor by the encoder.
 In this way, the auto-encoder method can construct descriptors of
 vectors from just a set of training vectors.

 However, a descriptor of an image from an ordinary auto-encoder
 includes
 both a spatial subpattern and its position.
 %
 If images have a common spatial subpattern at different positions,
 their descriptors are different.

 \section{TRANSFORM INVARIANT AUTO-ENCODER}
 As a method to construct a descriptor representing a property invariant
 to a certain set of transforms,
 we propose a transform invariant auto-encoder.
 We call the set ``ignored transforms''.
 A transform invariant auto-encoder is generated by training an
 auto-encoder with a novel cost function.
 The cost function should induce the accurate restoration of a
 subpattern as well as transform invariance.
 We achieve such an cost function by adding a transform variance
 term and relaxing the restoration error term.

 \subsection{Transform Variance Term}
 As a measure of the transform variance, we propose a sum of differences
 between a restored image and an image restored from a transformed input
 as follows:
 \begin{equation}
  \label{eq:3}
   C_{\text{inv}}(E, D)
   \defineequal
   \sum_{I \in S}
   \sum_{i}
   \norm{D\left(E\left(I\right)\right)-
   D\left(E\left(T_{\theta_{i}}\left(I\right)\right)\right) }^{2}_{2},
 \end{equation}
 where $S$ and $T_{\theta}$ denote a set of training inputs and a
 transform operator in the ignored transforms, respectively.
 To minimize (\ref{eq:3}), the combination of the encoder
 $E$ and the decoder $D$ need to output similar vectors for variously
 transformed versions of an input.
 By optimizing the encoder $E$ and the decoder $D$ so that they minimize
 (\ref{eq:3}), their combination is approximately transform invariant
 for inputs in the set $S$.
 %
 %

 %
 \subsection{Restoration Error Term}
 To compare subpatterns without respect to ignored transforms,
 we need to relax the restoration error cost in (\ref{eq:1})
 so that it will be small if a restored input matches a
 transformed version of its original input.
 Therefore, we propose the following term as a measure of the accuracy
 of the restoration of a subpattern:
 \begin{equation}
  \label{eq:4}
   C_{\text{res}}(E, D)
   \defineequal
  \sum_{I \in S}
  \min_{i} \norm{T_{\theta_i}
  \left(I\right) - D\left(E\left(I\right)\right)}^{2}_{2}.
 \end{equation}
 To minimize (\ref{eq:4}), the restored image $D(E(I))$
 should approximately match one of the transformed inputs
 $\{T_{\theta_i}(I)\}$.
 This means that the subpattern should be approximately
 restored.

 \subsection{Cost Function}
 Our total cost function $C(E, D)$ is formulated as follows;
 \begin{equation}
  \label{eq:2}
   \begin{aligned}
    C\left(E, D\right)
    \defineequal
    &
    \lambda_{\text{inv}} C_{\text{inv}}\left(E, D\right)
    +
    \lambda_{\text{res}} C_{\text{res}}\left(E, D\right)
    \\
    &
    +
    \lambda_{\text{spa}}
    \sum_{I \in S}
    \left(
    \frac
    {\norm{E\left(I\right)}_{1}}
    {\norm{E\left(I\right)}_{2}}
   \right)^{2},
   \end{aligned}
 \end{equation}
 where $\lambda_{\text{inv}}$,
 $\lambda_{\text{res}}$, and $\lambda_{\text{spa}}$
 denote the scalar weights of each term.
 The third term evaluates the spatial sparseness of the
 descriptors \cite{matsuo2016fcv}.

 We train the encoder $E$ and the decoder $D$ so that they minimize the
 proposed cost function $C\left(E, D\right)$.

 \section{INFERENCE OF TRANSFORM PARAMETER}
 We also propose a inference method of a transform parameter which is
 ignored by a transform invariant auto-encoder.
 We define a transform parameter of an input $I$ as a parameter
 representing a transform from the input $I$ to the restored input
 $D(E(I))$.
 For example, a transform parameter for a shift invariant auto-encoder
 means a spatial shift.
 An input can be approximately restored from its descriptor and
 transform parameter.
 Therefore, a pair of a transform invariant auto-encoder and
 the corresponding inference model of a transform parameter is an
 auto-encoder that can represent an input as a pair of a transform
 invariant part and a transform variant part.

 We propose the following cost function to train an inference model $R$
 of a transform parameter.
 \begin{equation}
  C_{\text{par}}(R)
   =\sum_{I \in S}
   \norm{
   R\left(I\right)
   - \argmin_{\theta}
   \norm{I -
   T_{\theta}\left(D\left(E\left(I\right)\right)\right)}^{2}_{2}
   }^{2}_{2}.
 \end{equation}
 We can achieve an inference model $R$ of a transform parameter by
 minimizing $C_{\text{par}}(R)$.

 \section{EXPERIMENTS}
 We demonstrate the effectiveness of the proposed method
 by experiments with a shift invariant auto-encoder.
 The shift operator $T_{\theta_i}$ is defined as
 \begin{equation}
  \left(T_{\theta_i}\left(I\right)\right)(x, y)
   = I(x + \Delta x_{i}, y + \Delta y_{i}),
 \end{equation}
 where $I(x,y)$ denotes the value of the image $I$ at the position
 $(x, y)$.
 We used the following shift
 parameters:
 \begin{equation}
  \label{eq:5}
  \left\{\left(\Delta x_{i}, \Delta y_{i}\right)\right\}
   = \left\{-8, -6, -4, -2, 0, 2, 4, 6, 8\right\}^{2}.
 \end{equation}

 \subsection{Experiments for MNIST}
 \label{sec:experiments-digits}
 Here, we demonstrate shift invariant property of the proposed method
 using
 experiments for digit patterns.

 As an encoder, we used a neural network consisting of
 a single CNN with
 $9\times 9$ filter kernels and 16-channel outputs following a max
 pooling with stride 2 and a three-layer fully connected neural network
 (NN), where each layer has 1500, 150, 30 outputs respectively.
 As a decoder, we used a three-layer fully connected NN, where each
 layer has 150, 1500, 1024 outputs, respectively.
 In addition, we used a hyperbolic tangent as an activation
 function, which is placed between each pair of layers.
 We generated two pairs of encoders and decoders with the same
 structure.
 One was trained as an ordinary auto-encoder by minimizing (\ref{eq:1}),
 and the other was trained as a shift invariant auto-encoder by
 minimizing (\ref{eq:2})
 for digit images of training images in the MNIST database
 \cite{726791}.
 For the ordinary auto-encoder, we used additional images that were
 randomly shifted according to the parameters in (\ref{eq:5}).
 Both auto-encoders were trained by stochastic gradient descent (SGD)
 \cite{726791} with learning rate $1.0\times 10^{-3}$, and both
 were updated with every 50 samples that were randomly extracted from
 the training images (60k samples) in the MNIST database.
 We used auto-encoders that were updated 100,000 times ($\approx$ 83 epochs).
 We also trained an inference model $R$ of a shift parameter.
 The inference model consisted of a three-layer fully connected NN.

\begin{figure*}[t]
  \vspace{.4em}
   \begin{center}
    \begin{minipage}[b]{.2\textwidth}
     {\centering
     \includegraphics[width=0.9\textwidth]{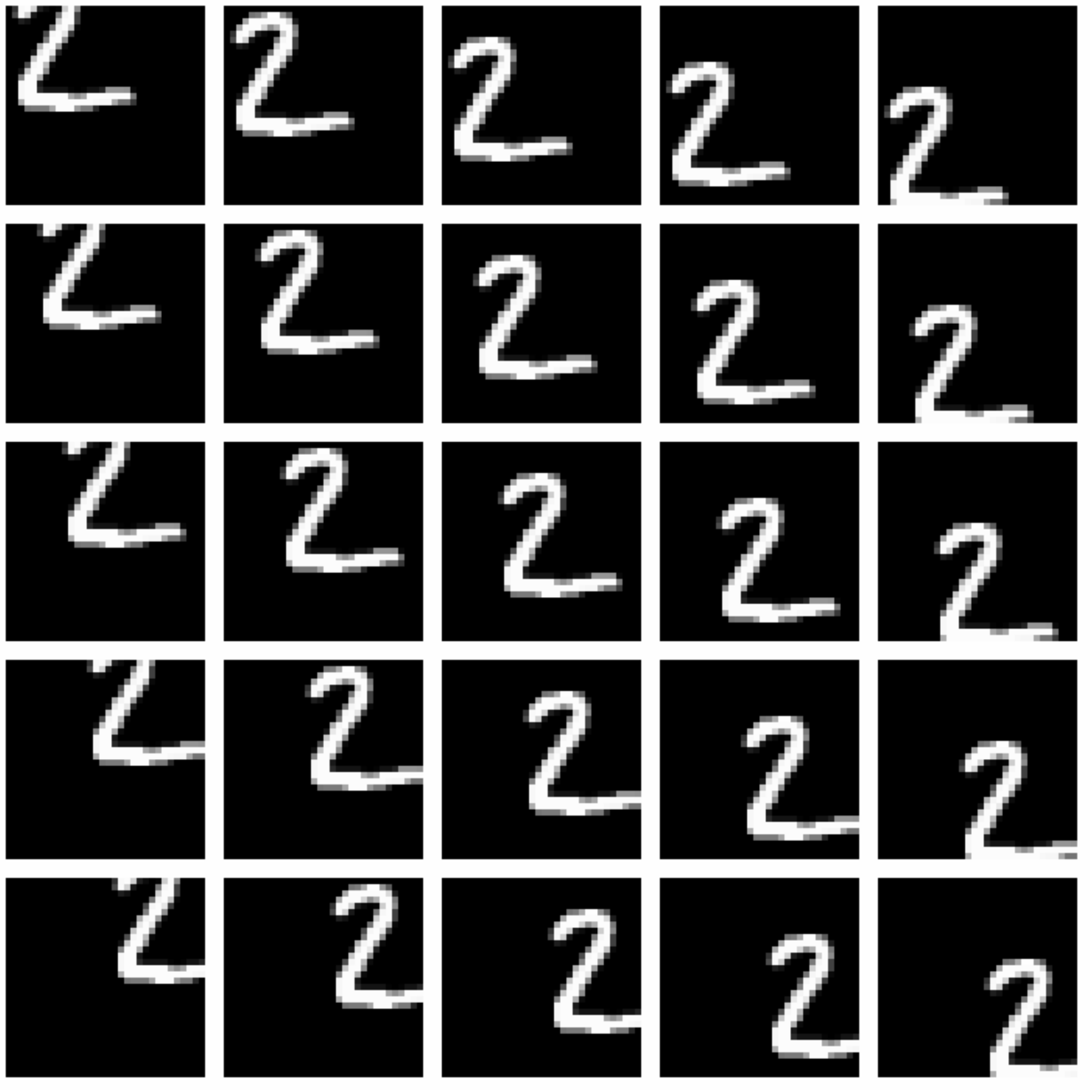}
     \caption{An image in MNIST and its shifted versions}
     \label{fig:mnist_shifted_input}
     }
    \end{minipage}
    \begin{minipage}[b]{.2\textwidth}
     {\centering
     \includegraphics[width=0.9\textwidth]{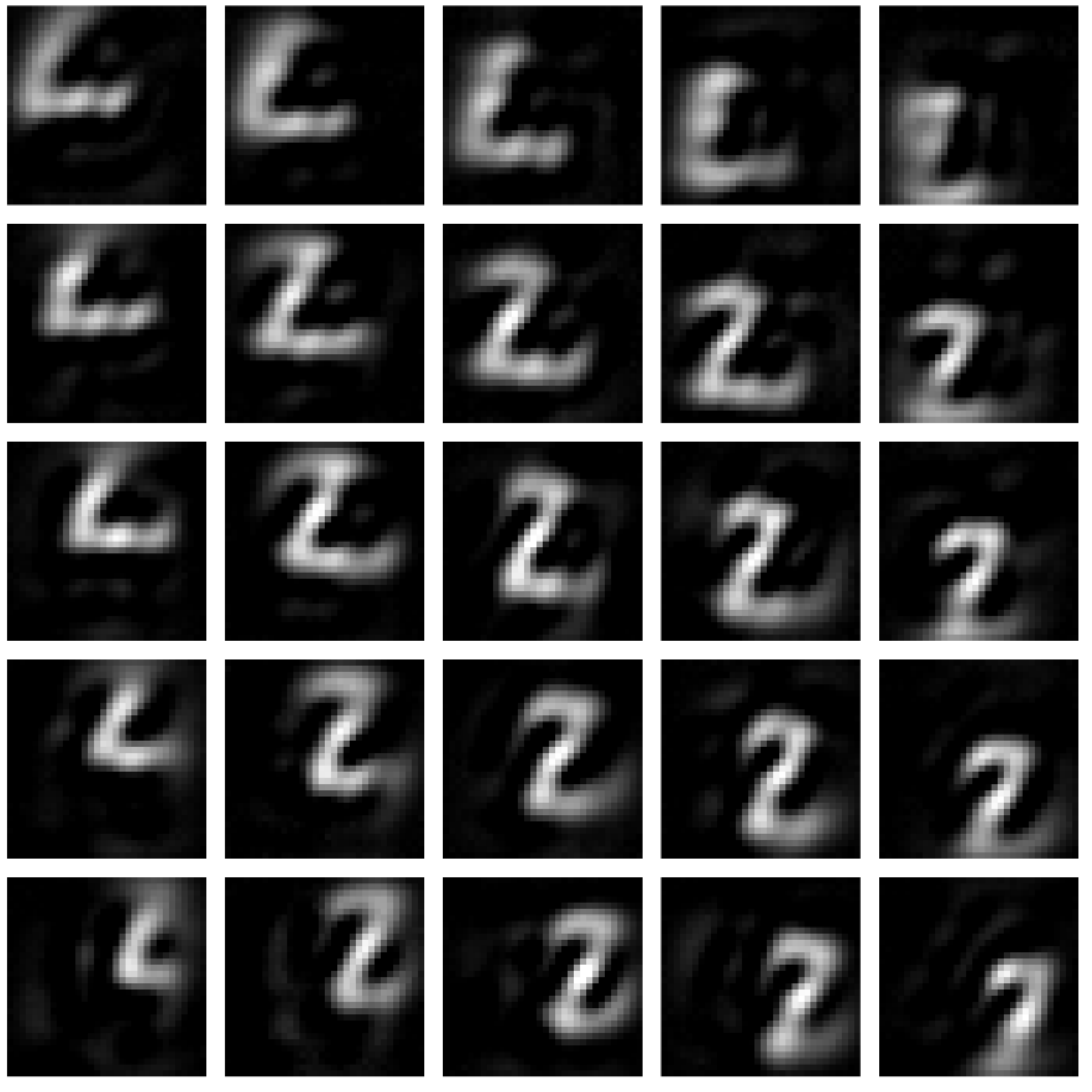}
     \caption{Images restored using an ordinary auto-encoder}
     \label{fig:mnist_ordinary_ae_restored}
     }
    \end{minipage}
    \begin{minipage}[b]{.2\textwidth}
     {\centering
     \includegraphics[width=0.9\textwidth]{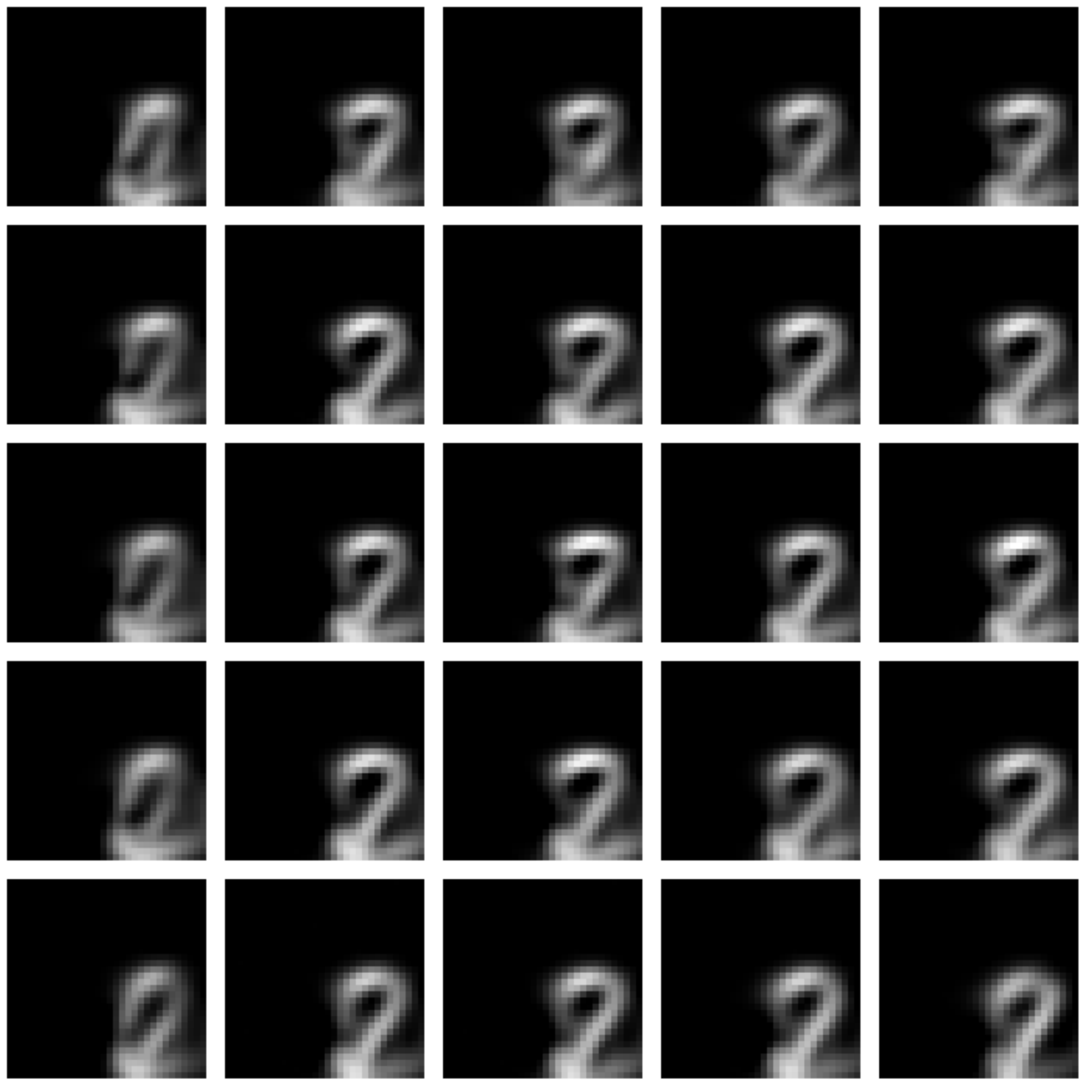}
     \caption{Images restored using a shift invariant auto-encoder}
     \label{fig:mnist_shift_invariant_ae_restored}
     }
    \end{minipage}
    \begin{minipage}[b]{.2\textwidth}
     {\centering
     \includegraphics[width=0.9\textwidth]{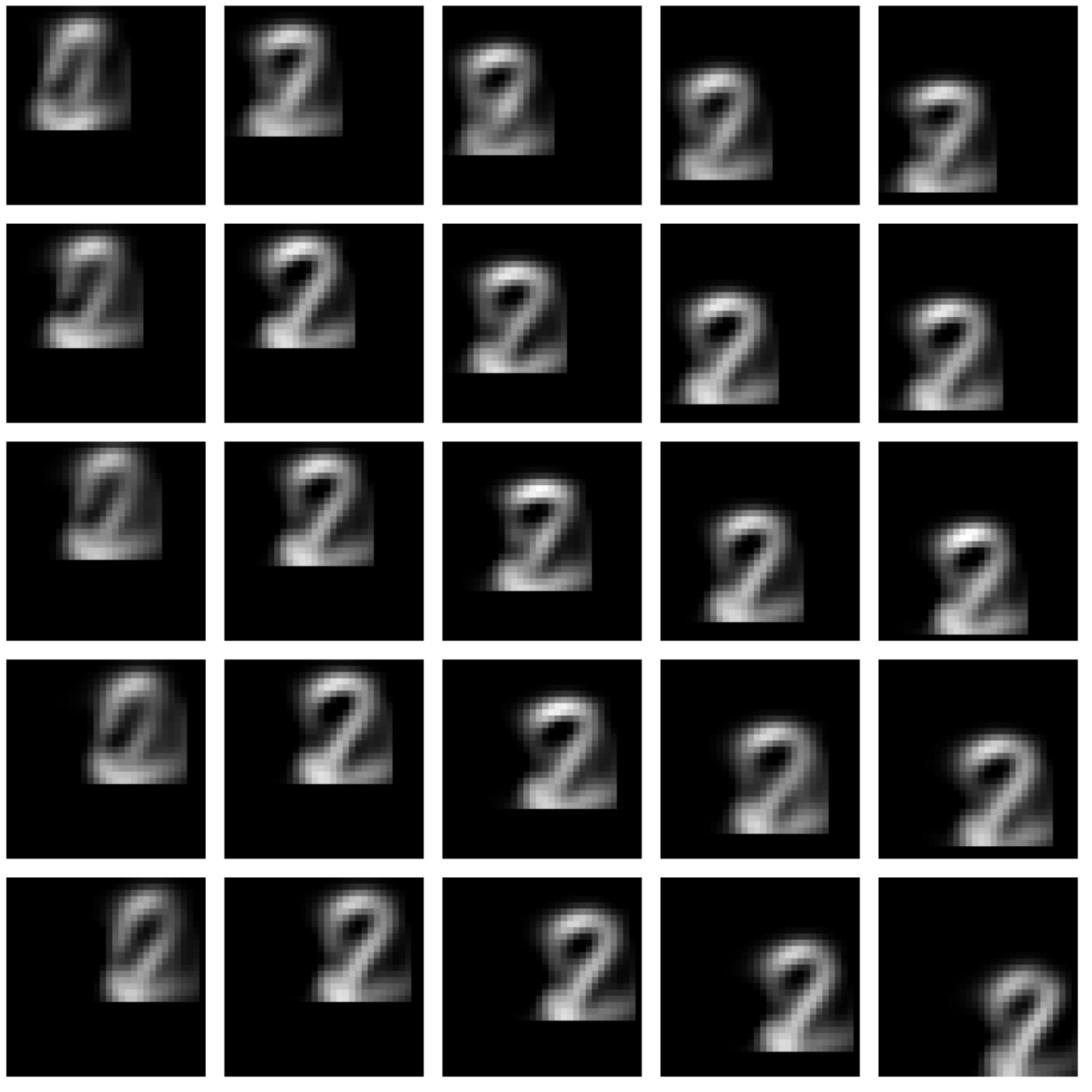}
     \caption{Images restored using a shift invariant auto-encoder with inferred shifts}
     \label{fig:mnist_shift_invariant_ae_restored_with_shift}
     }
    \end{minipage}
   \end{center}
 \end{figure*}

 As an example, we encoded and decoded an test image of the digit ``2'',
 which is not used in training auto-encoders.
 Input images are shown in Fig.~\ref{fig:mnist_shifted_input},
 where the center image is the original image in the MNIST database and
 the others are its shifted versions.
 Images in Fig.~\ref{fig:mnist_ordinary_ae_restored} are restored from
 images in Fig.~\ref{fig:mnist_shifted_input} using an ordinary
 auto-encoder.
 Images restored by a proposed shift invariant auto-encoder are shown in
 Fig.~\ref{fig:mnist_shift_invariant_ae_restored}.
 Fig.~\ref{fig:mnist_shift_invariant_ae_restored_with_shift} shows the
 restored images which are shifted according to the shift parameters
 estimated by the inference model $R$.
 In Fig.~\ref{fig:mnist_ordinary_ae_restored}, the restored images are
 located depending on the shifts in the input images.
 Conversely, the restored images in
 Fig.~\ref{fig:mnist_shift_invariant_ae_restored}
 are very similar to each other and they are closer to a typical shape
 of the digit ``2'' than the input in
 Fig.~\ref{fig:mnist_shifted_input}.
 In the cost function (\ref{eq:2}), there is a trade-off between
 $C_{\text{inv}}$ and $C_{\text{res}}$.
 In this case, the auto-encoder successfully find a typical shape of
 the digit ``2'' by focusing on shapes without their positions.

\begin{figure}[t]
    \begin{minipage}[b]{.2025\textwidth}
     \includegraphics[width=0.99\textwidth]{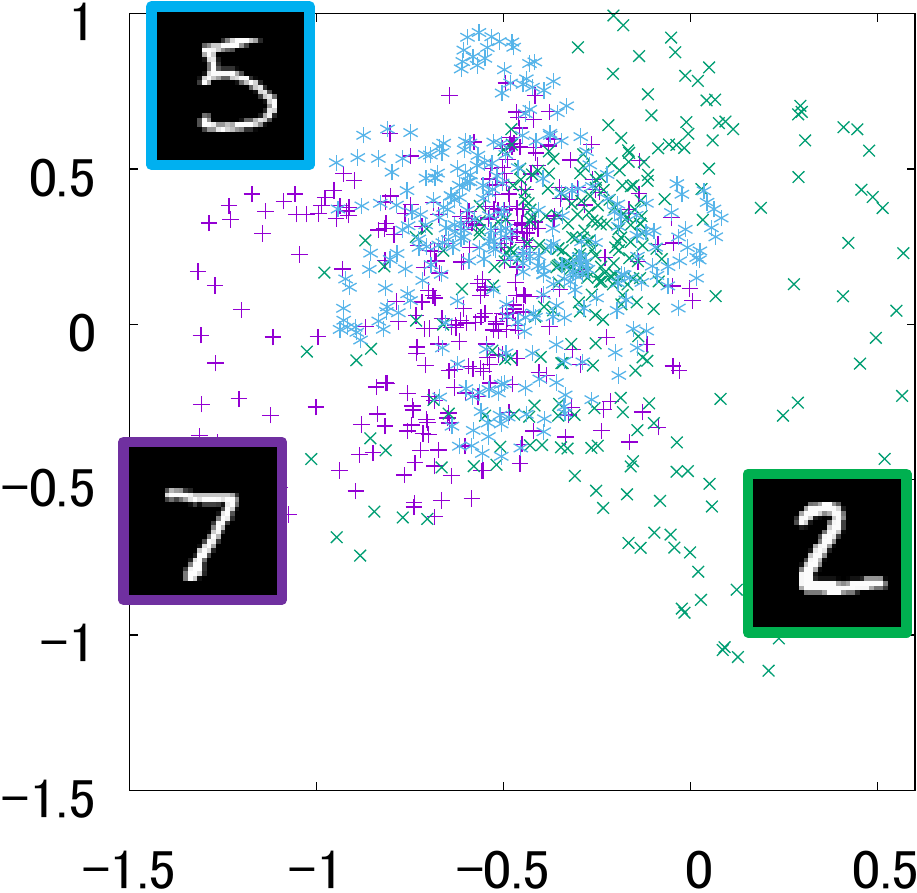}
     \caption{The distribution of descriptors of shifted images generated by an ordinary auto-encoder}
     \label{fig:distribution_of_desc_by_ordinary_auto_encoder}
    \end{minipage}
    \begin{minipage}[b]{.216\textwidth}
     \includegraphics[width=0.99\textwidth]{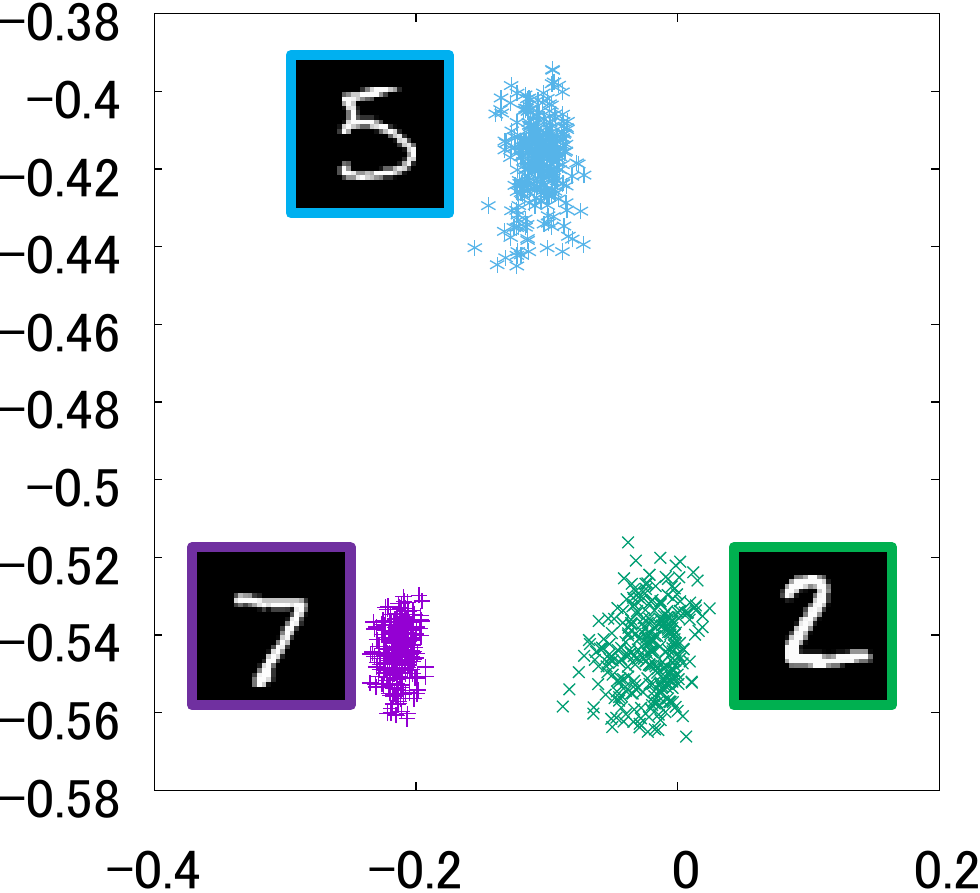}
     \caption{The distribution of descriptors of shifted images generated by an shift invariant auto-encoder}
     \label{fig:distribution_of_desc_by_shift_invariant_auto_encoder}
    \end{minipage}
  {\centering
   \includegraphics[width=0.33\textwidth]{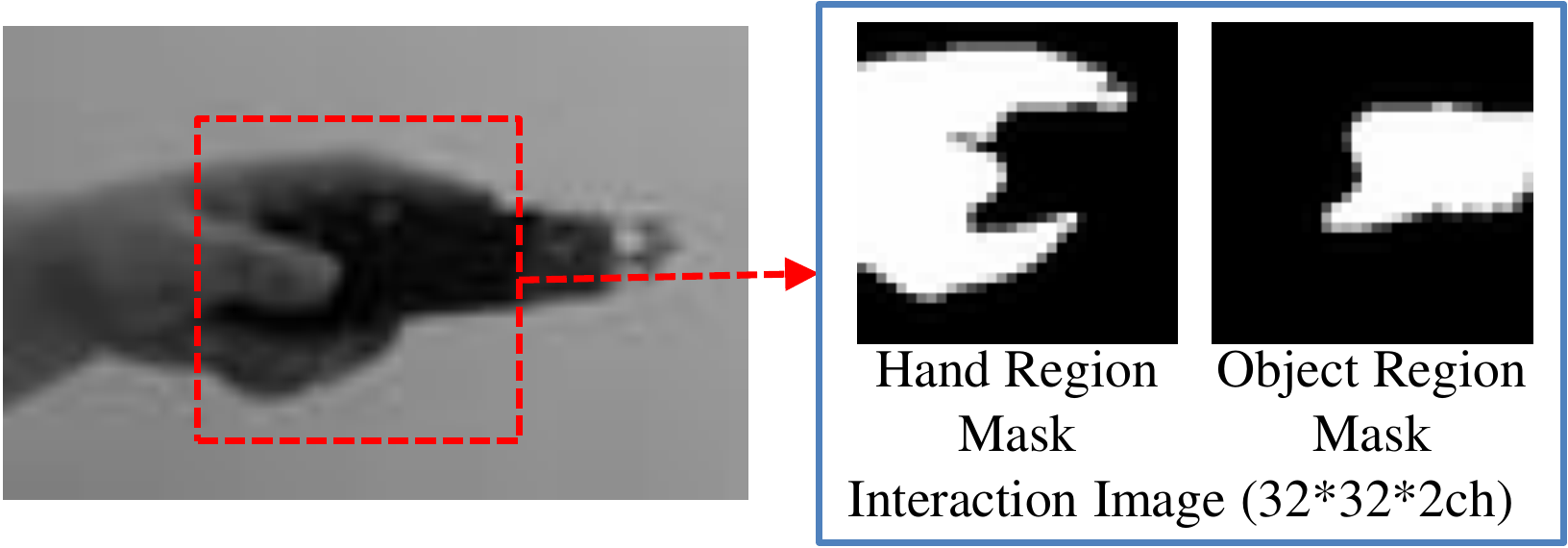}
   \caption{An interaction image}
   \label{fig:interaction_image}
  }
 \end{figure}

 In addition, we calculated the distributions of the descriptors from
 the shifted
 images.
 We encoded the digit images corresponding to ``2'', ``5'', and ``7''
 and
 their shifted versions using the two auto-encoders.
 Fig.~\ref{fig:distribution_of_desc_by_ordinary_auto_encoder} shows
 the distributions from the ordinary auto-encoder, and
 Fig.~\ref{fig:distribution_of_desc_by_shift_invariant_auto_encoder}
 shows those from the shift invariant auto-encoder.
 In these figures, 30 dimensional descriptors are projected onto a
 two-dimensional space spanned by the three mean vectors of the
 descriptors for digits ``2'', ``5'', and ``7''.
 By comparing these figures, we see that descriptors generated
 by the shift invariant
 auto-encoder are obviously concentrated for each digit.
 With a shift invariant auto-encoder,
 descriptors from shifted images of the same digit are close to each
 other
 and
 descriptors from shifted images of other digits are far from each
 other.
 This means that a descriptor generated by a shift invariant
 auto-encoder
 represents the spatial subpattern.
 In addition, descriptors in
 Fig.~\ref{fig:distribution_of_desc_by_shift_invariant_auto_encoder}
 make clusters corresponding to digits,
 even though we have entered no digit information when training the
 shift
 invariant auto-encoder.
 The proposed method may be applicable to the unsupervised clustering of
 images based on their spatial subpatterns.

 \subsection{Experiments for Hand Object Interactions}
 Here,
 we show examples of the encoding appearances of hand--object
 interactions,
 which are generally difficult to label or normalize.

\begin{figure}[t]
  {\centering
   \includegraphics[width=0.42\textwidth]{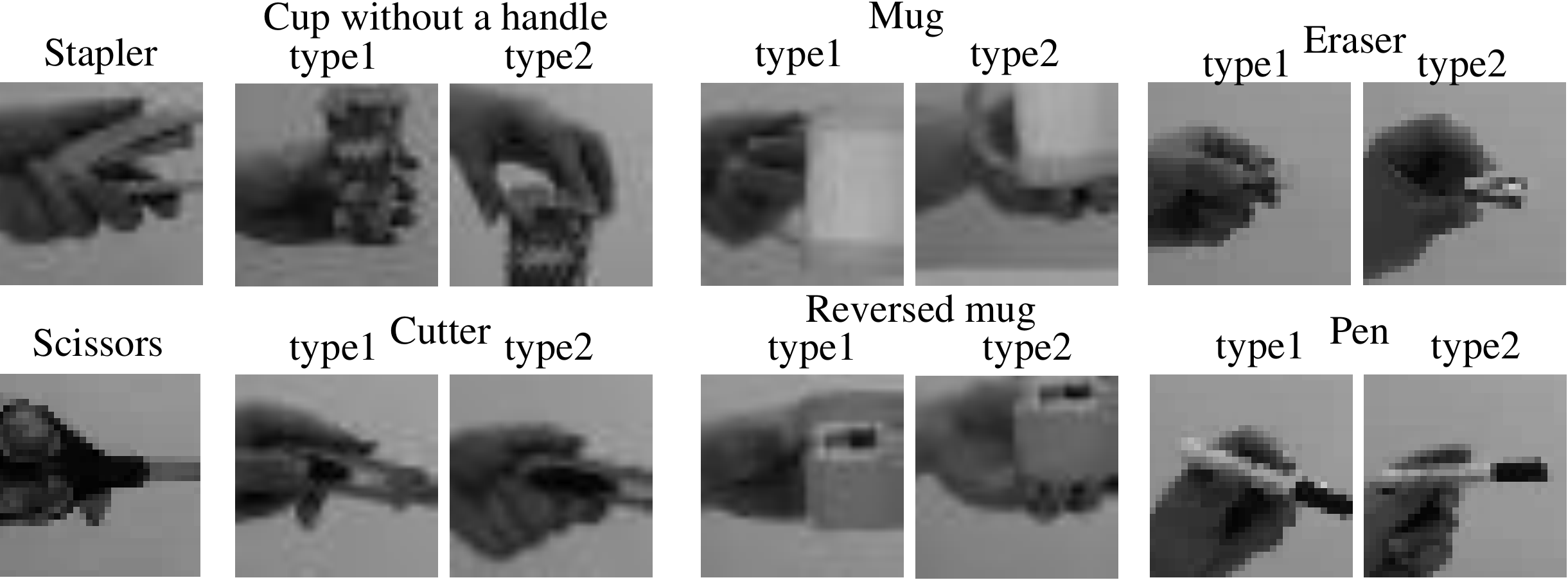}
   \caption{Interaction types}
   \label{fig:interaction_type}
  }
  {\centering
  \begin{minipage}[t]{0.15\textwidth}
   \includegraphics[width=0.99\textwidth]{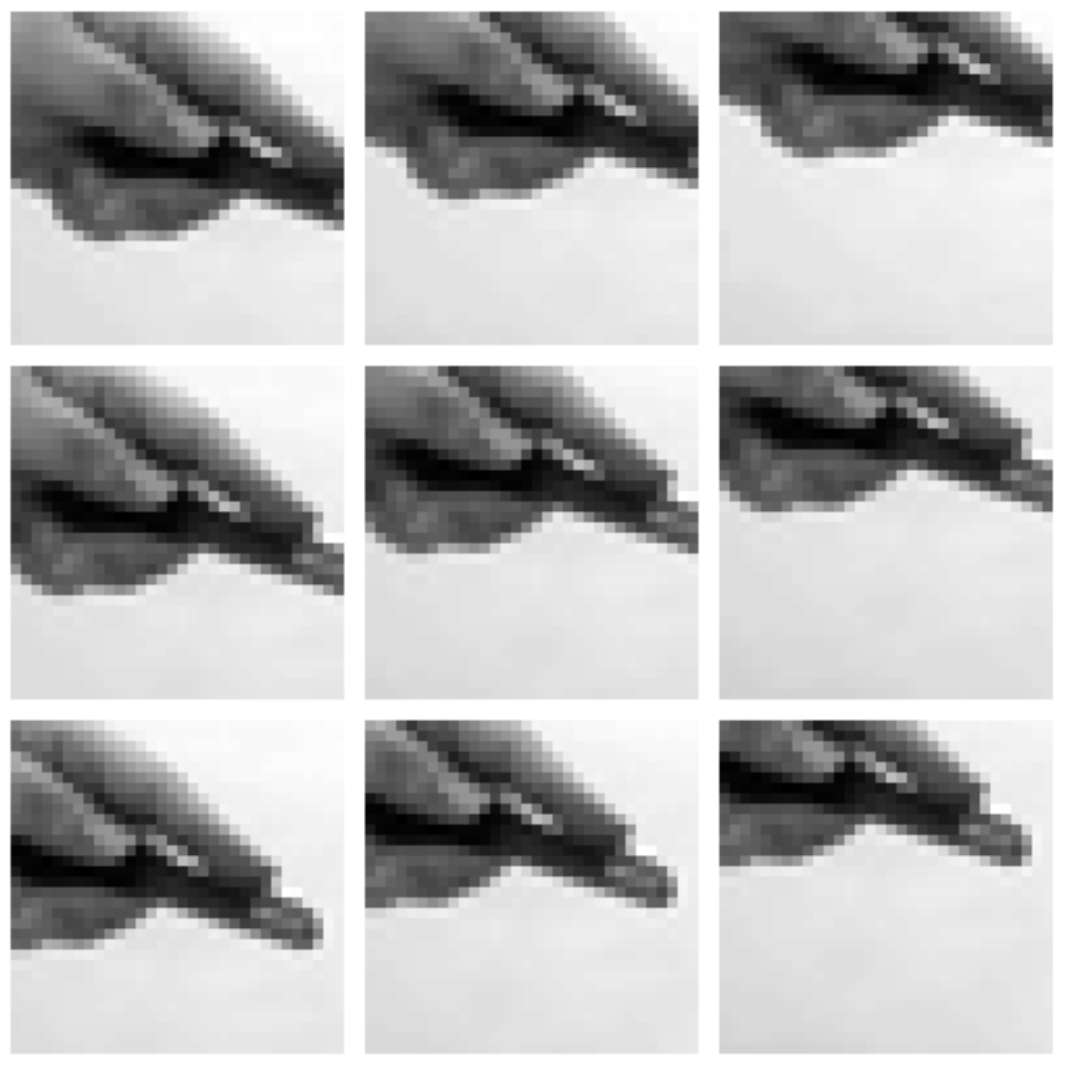}
   {\footnotesize (a) Appearances}
  \end{minipage}
  \begin{minipage}[t]{0.15\textwidth}
   \includegraphics[width=0.99\textwidth]{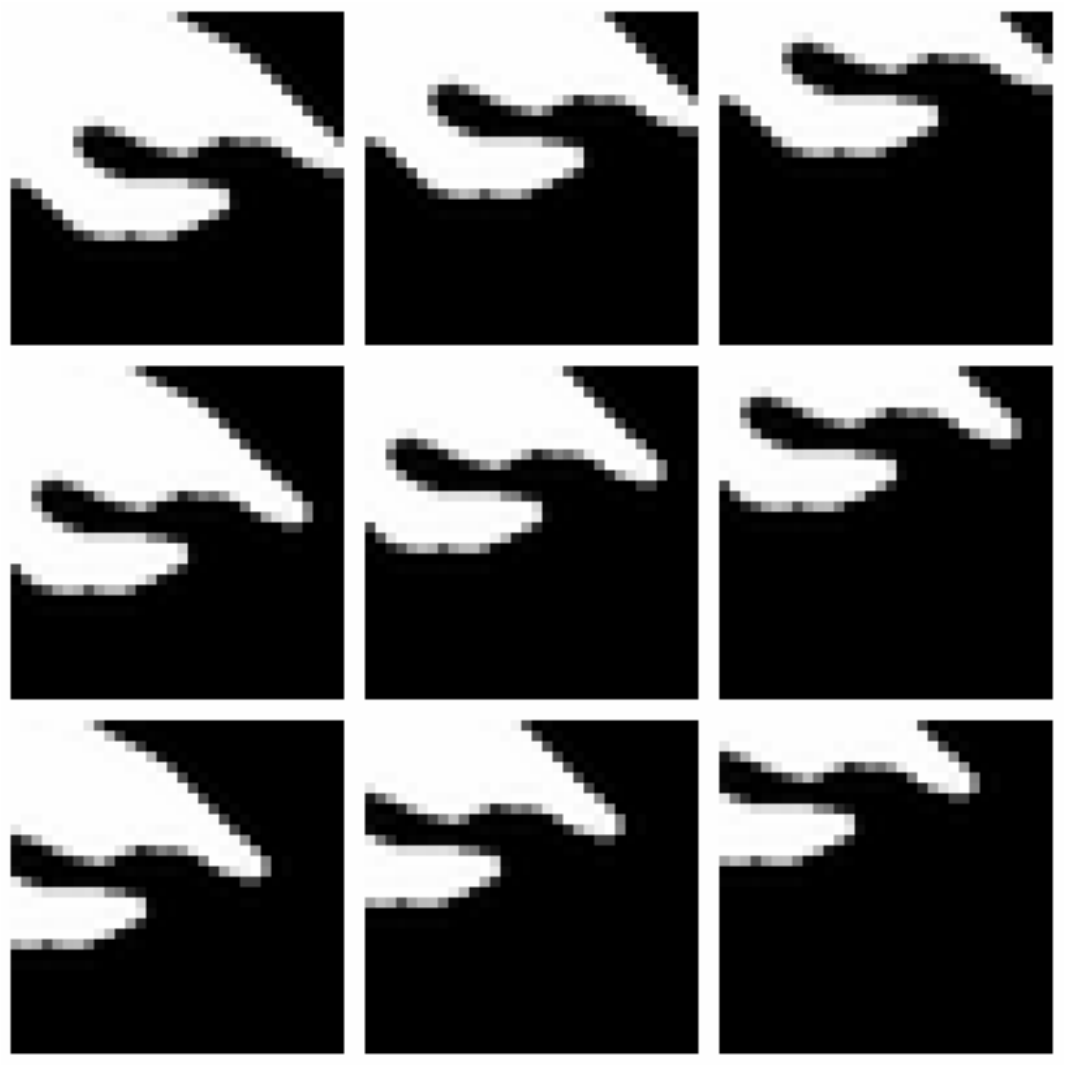}
   {\footnotesize (b) Hand region masks}
  \end{minipage}
  \begin{minipage}[t]{0.15\textwidth}
   \includegraphics[width=0.99\textwidth]{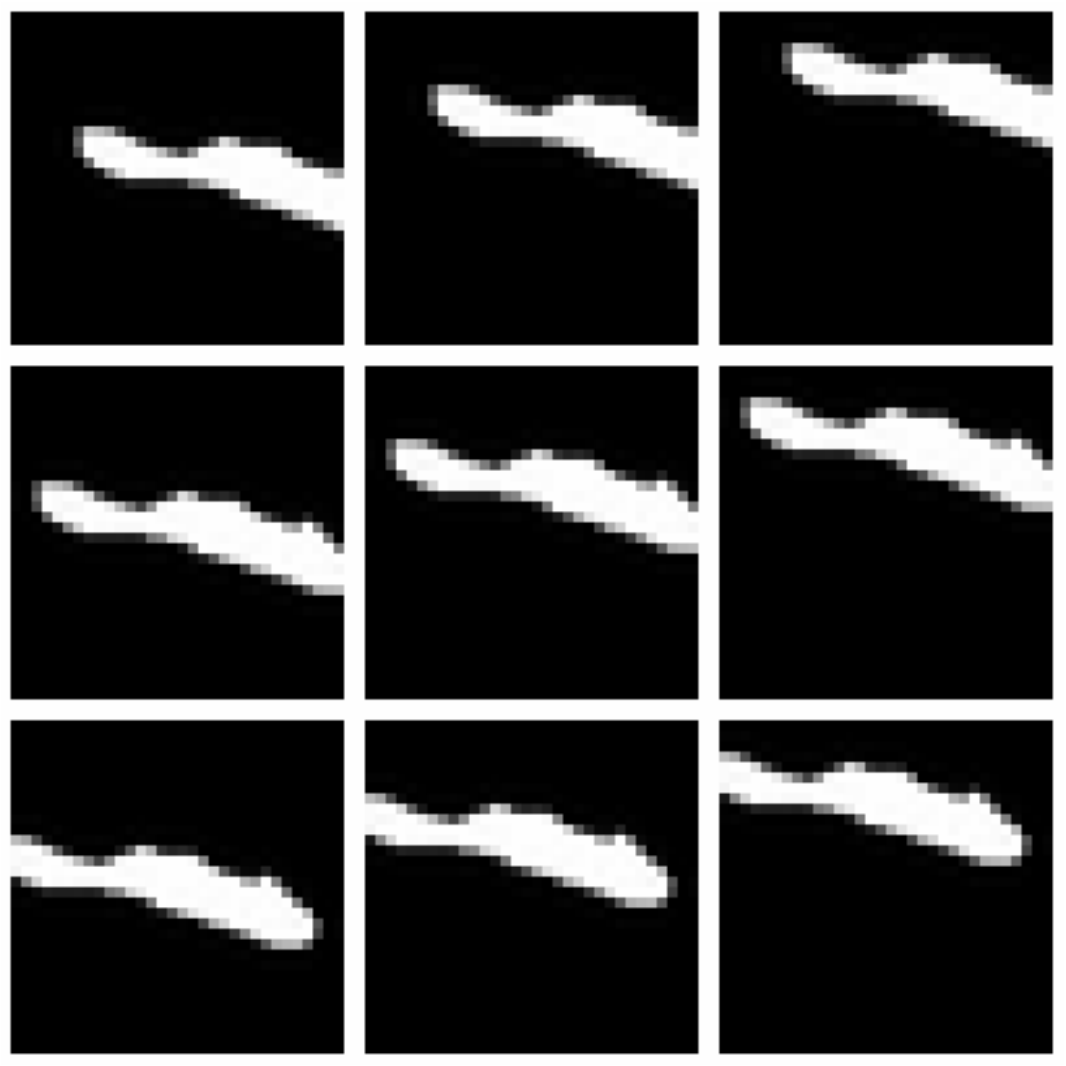}
   {\footnotesize (c) Object region masks}
  \end{minipage}
  \caption{Example of input interaction images}
  \label{fig:example_of_input_interaction_images}
  }
 \end{figure}

 In this experiment, we used a two-channel image consisting of a
 hand region mask and an object region mask
 (Fig.~\ref{fig:interaction_image}) as the input of the auto-encoders.
 The structures of the encoder and the decoder
 are similar to those used in \ref{sec:experiments-digits} except
 that the input of the encoder and the output of the decoder are
 two-channel $(32\times 32) [\text{pixel}]$ interaction images.
 We generated interaction images from the interaction scene images with
 a
 simple background via skin color extraction and background subtraction.
 We trained an ordinary auto-encoder and a shift invariant
 auto-encoder with interaction images that included a hand region larger
 than 20\% of the entire image.
 The interaction images were extracted from random positions of 1680
 scenes
 including the 14 types of interactions shown in
 Fig.~\ref{fig:interaction_type}.
 Both auto-encoders were trained using SGD and
 both were updated with every 168 samples randomly extracted from
 the 1680 scenes.
 We used auto-encoders updated 20,000 times.


\begin{figure}[t]
  \vspace{.4em}
  {\centering
   \begin{tabular}{cc}
    \includegraphics[width=0.15\textwidth]{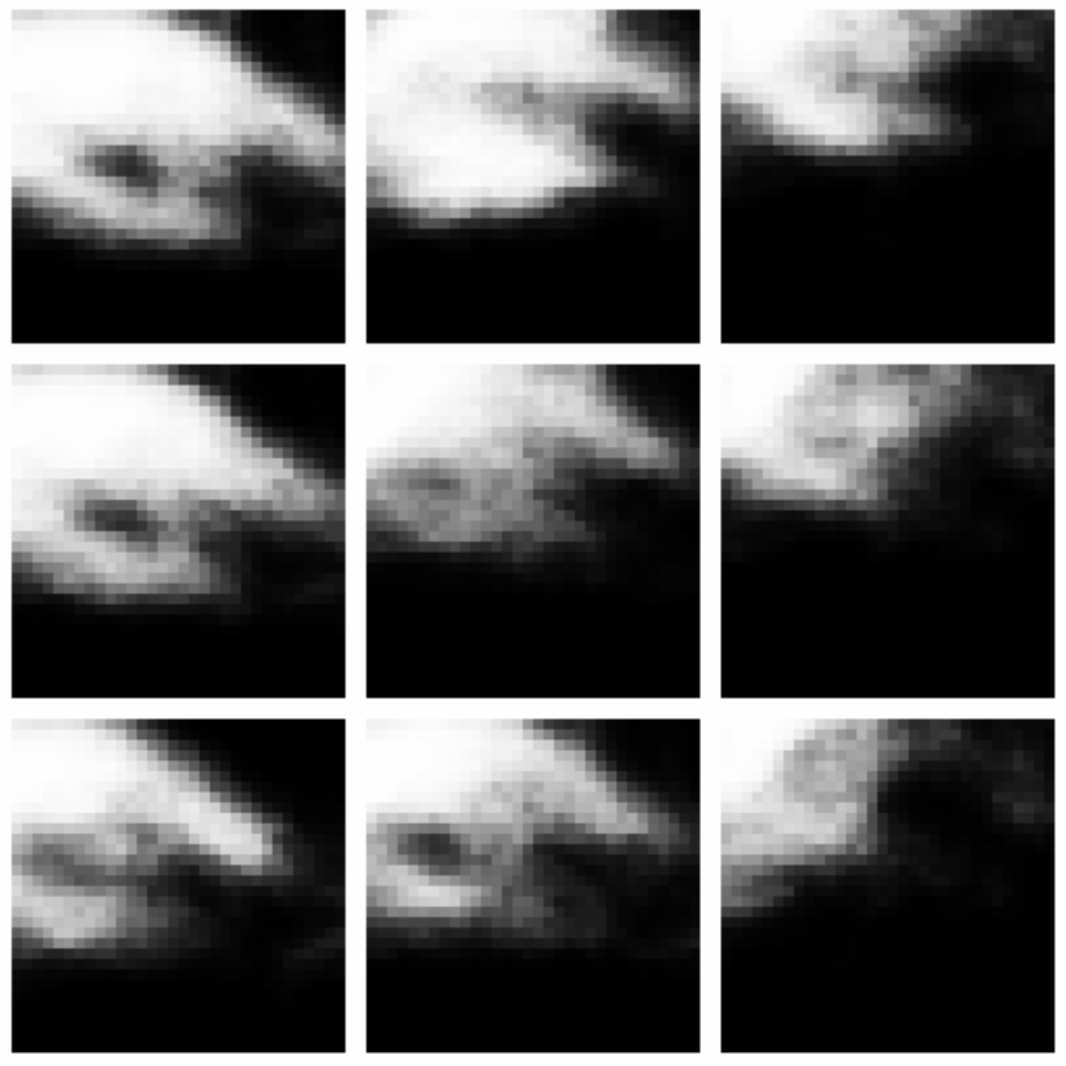}
    &
    \includegraphics[width=0.15\textwidth]{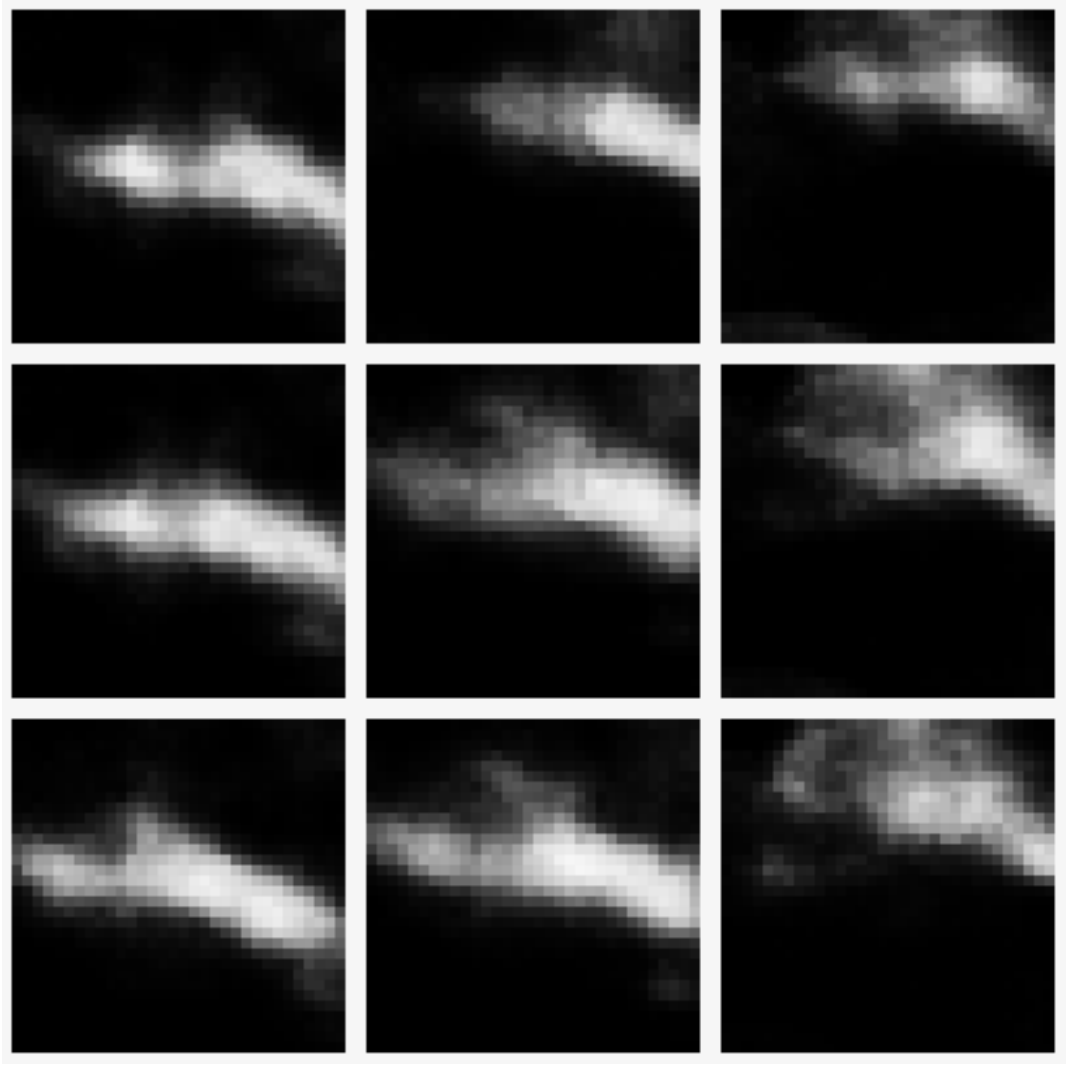}
    \\
    {\footnotesize (a)Hand region masks}&
    {\footnotesize (b)Object region masks}
   \end{tabular}
  \caption{Interaction images restored using an ordinary auto-encoder}
  \label{fig:interaction_images_restored_by_ordinary_ae}
   \begin{tabular}{cc}
    \includegraphics[width=0.15\textwidth]{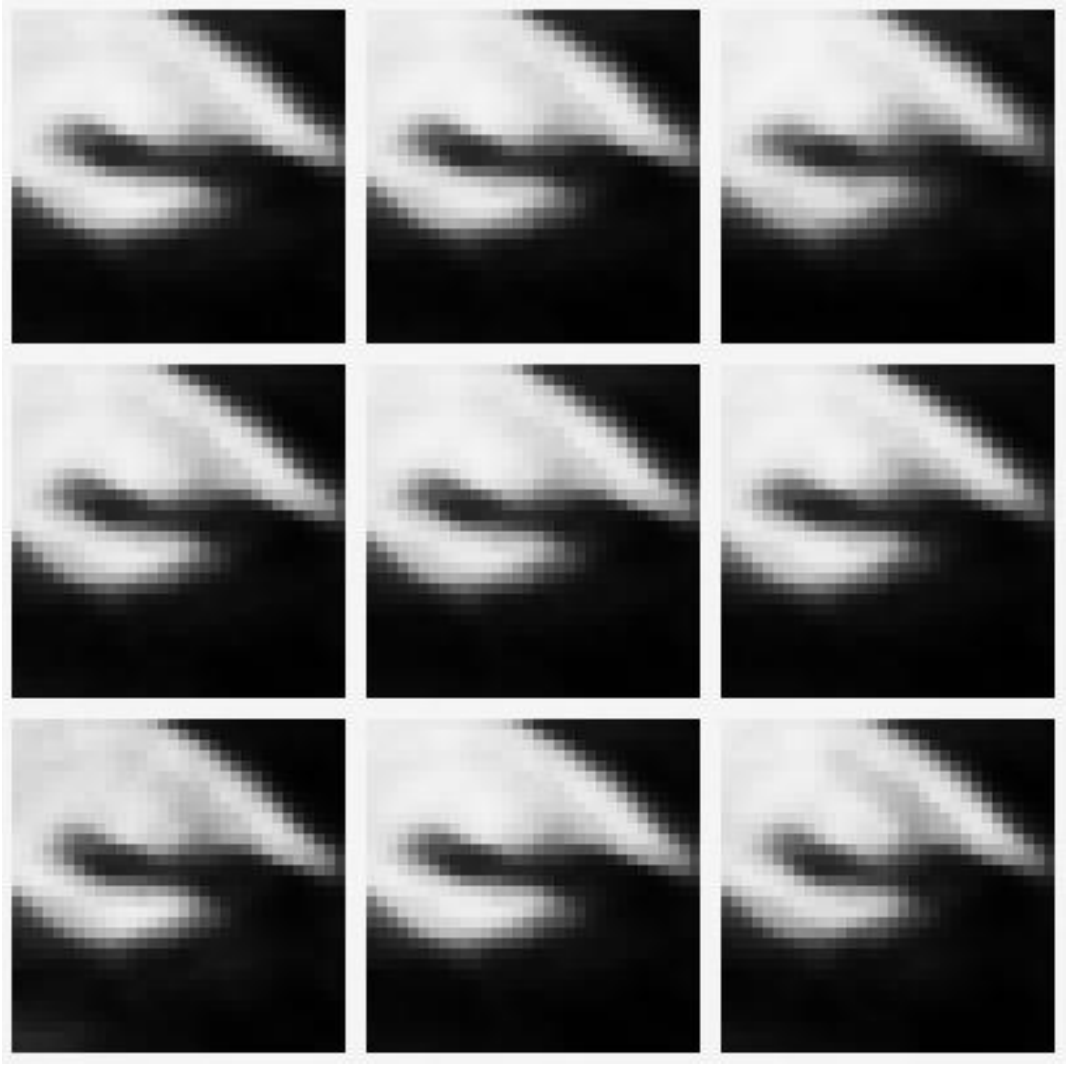}
    &
    \includegraphics[width=0.15\textwidth]{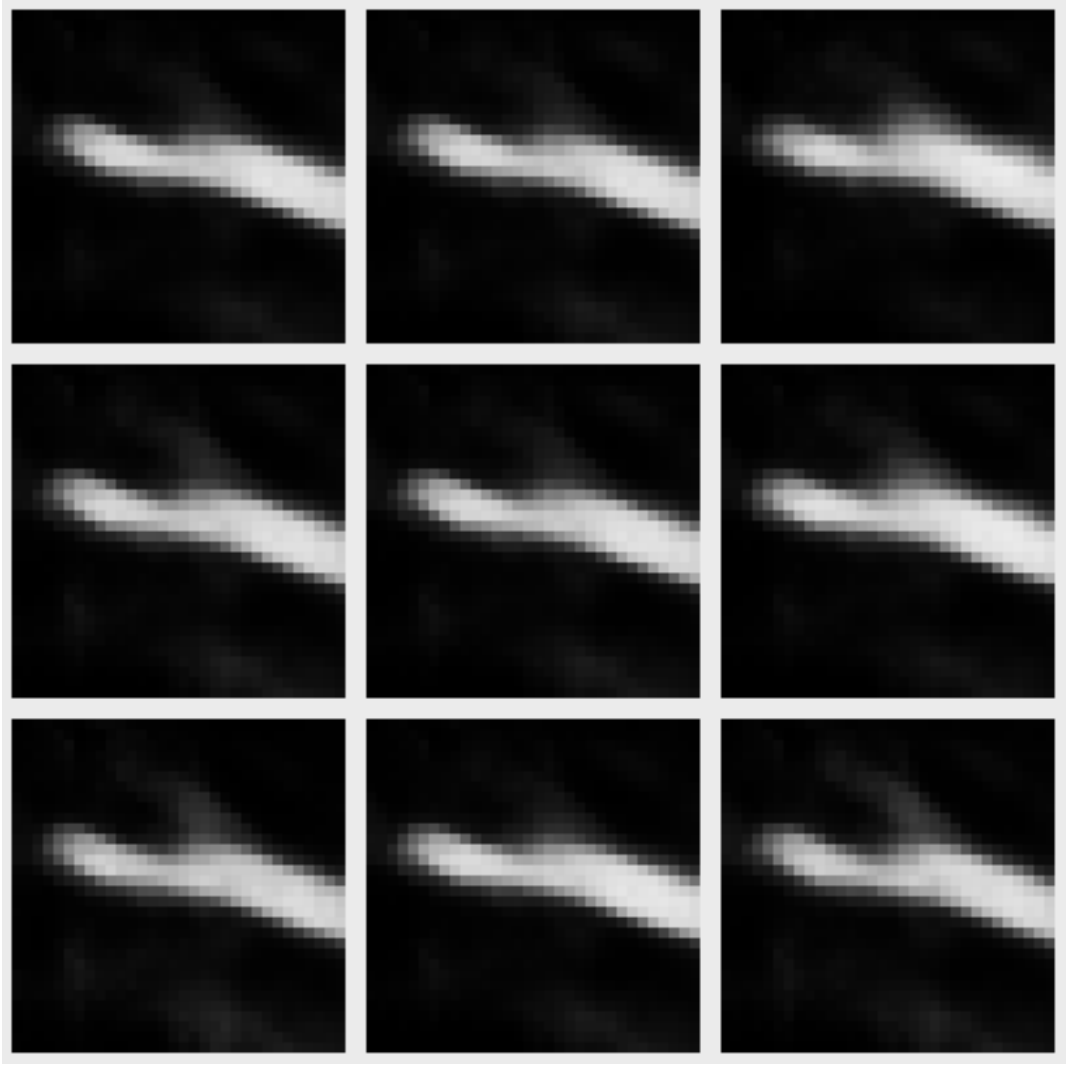}
    \\
    {\footnotesize (a)Hand region masks}&
    {\footnotesize (b)Object region masks}
   \end{tabular}
  \caption{Interaction images restored using a shift invariant auto-encoder}
  \label{fig:interaction_images_restored_by_shift_invariant_ae}
  }
 \end{figure}

 To compare the spatial patterns represented by the descriptors using an
 ordinary
 auto-encoder and a shift invariant auto-encoder, we encoded and
 decoded the interaction images used in the training process.
 Fig.~\ref{fig:example_of_input_interaction_images} shows nine
 interaction images extracted from a scene where a cutter is grasped.
 Fig.~\ref{fig:interaction_images_restored_by_ordinary_ae} shows the
 interaction images restored using the ordinary auto-encoder.
 In the figure, outlines of the restored hand region masks are blurred
 and
 differ depending on the shifts in the input images.
 In particular, the restored images in the right column are very
 different
 from their input images.
 Fig.~\ref{fig:interaction_images_restored_by_shift_invariant_ae} shows
 the interaction images restored using the shift invariant auto-encoder.
 Contrary to the ordinary auto-encoder,
 the interaction images restored by the shift invariant auto-encoder
 have clearer outlines.
 This shows that the descriptors generated by the shift invariant
 auto-encoder
 represent spatial subpatterns more accurately than those generated by
 an
 ordinary auto-encoder, as we expected.

 The restored images are similar to the center image in the left column
 of Fig.~\ref{fig:example_of_input_interaction_images}.
 This means that the image is considered to be a typical image in the
 training process of the shift invariant auto-encoder.
 In addition,
 the bottom image in the right column of
 Fig.~\ref{fig:interaction_images_restored_by_shift_invariant_ae}
 is similar to the typical image
 even though the corresponding input image in
 Fig.~\ref{fig:example_of_input_interaction_images} includes only
 fingertips.
 This means that a shift invariant auto-encoder can predict a possible
 neighbor typical pattern from a local non-typical pattern.
 The shift invariant auto-encoder enables us to analyze an image using
 typical features without a dense scan.

\begin{figure*}[t]
  \vspace{.4em}
  \begin{minipage}[c]{0.75\textwidth}
  {\centering
  \begin{tabular}{cc}
   \includegraphics[width=0.45\textwidth]{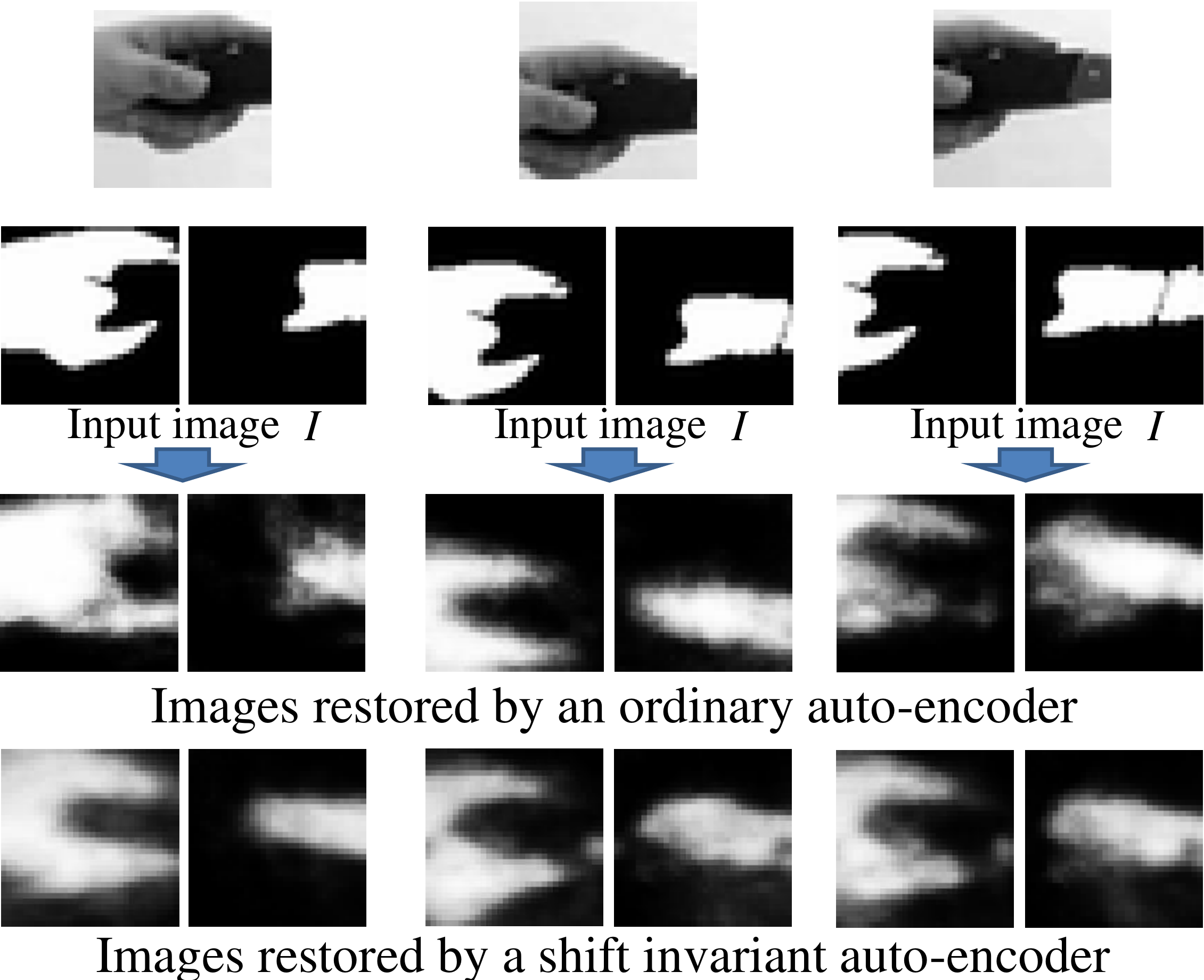}
   &
   \includegraphics[width=0.45\textwidth]{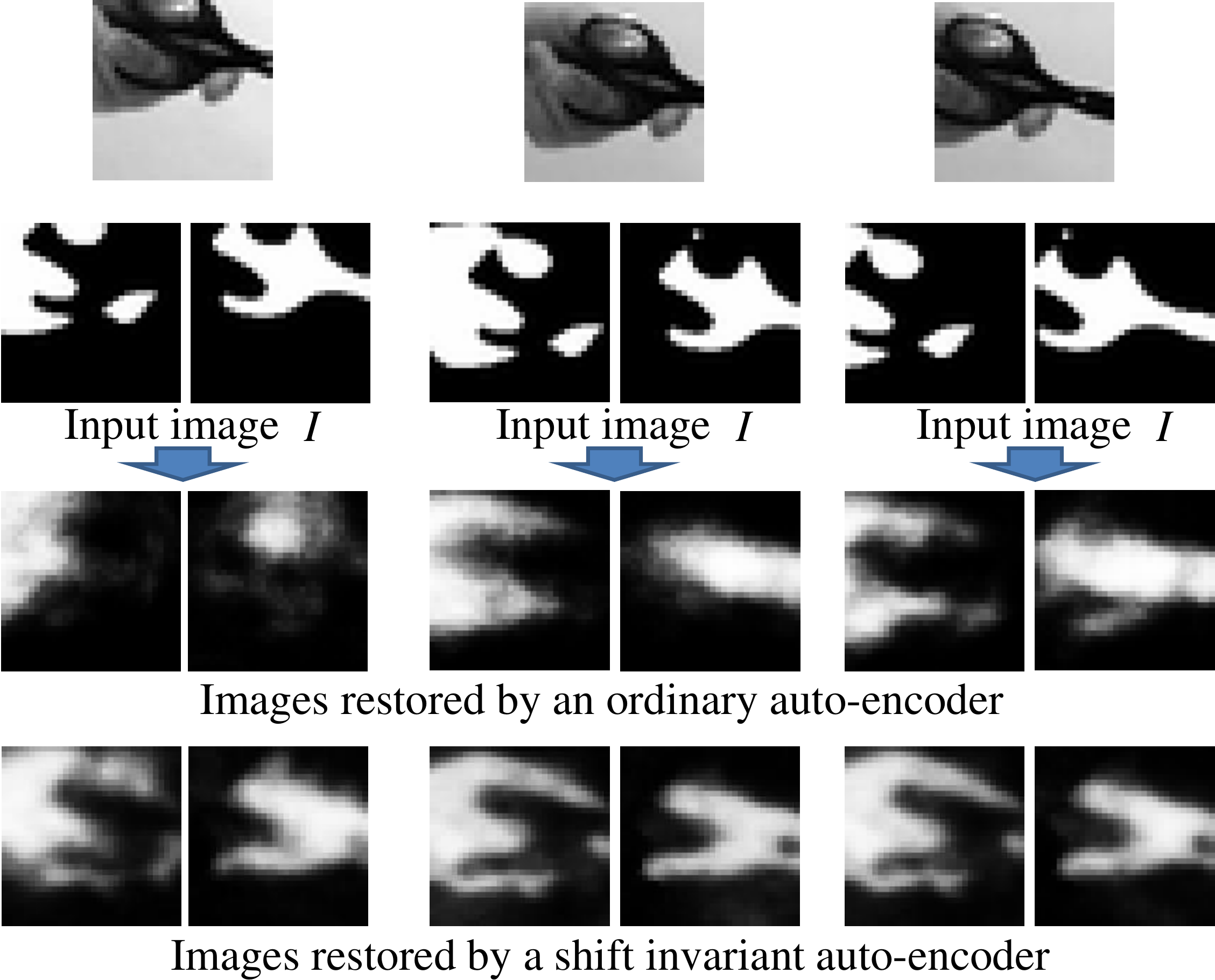}
   \\
   {\footnotesize (a) Unknown interaction images with a cutter}
   &
   {\footnotesize (b) Unknown interaction images with scissors}
  \end{tabular}
  \caption{Restoration from unknown interaction images}
  \label{fig:restoration_from_unknown_interaction_images}
  }
  \end{minipage}
  \begin{minipage}[c]{.12\textwidth}
   {\centering
   \includegraphics[width=.99\textwidth]{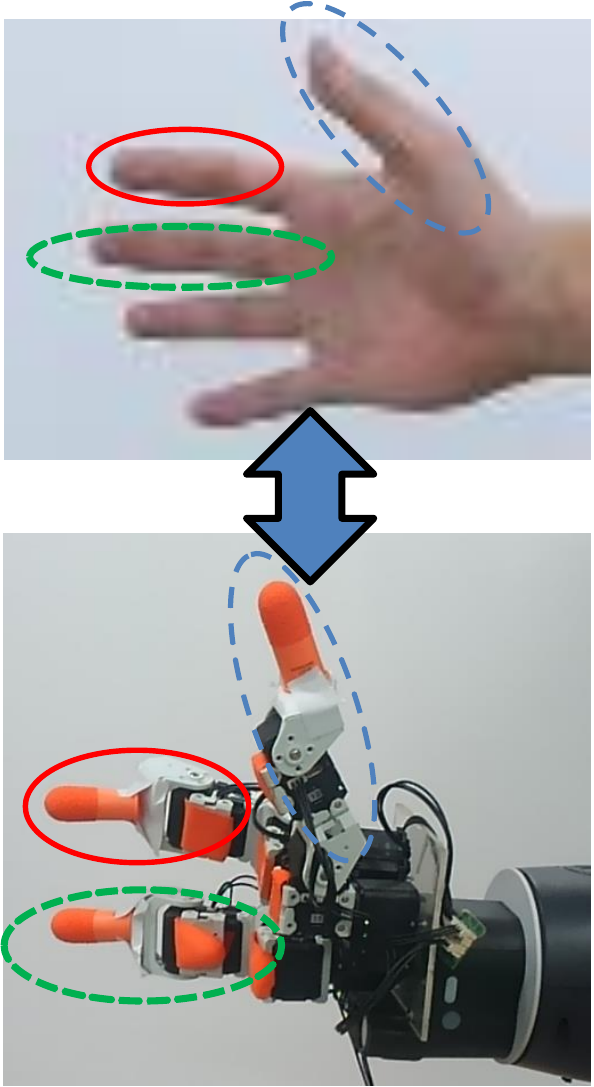}
   \caption{Correspondence of the human hand to the robot hand}
   \label{fig:correspondence_of_fingers}
   }
  \end{minipage}
 \end{figure*}

 In addition,
 we applied similar experiments to unknown interaction images
 that are not used in the training process.
 Fig.~\ref{fig:restoration_from_unknown_interaction_images} shows
 interaction images restored by the ordinary auto-encoder and the shift
 invariant auto-encoder.
 Fig.~\ref{fig:restoration_from_unknown_interaction_images}(a) is the
 case of grasping a cutter, and
 Fig.~\ref{fig:restoration_from_unknown_interaction_images}(b) is the
 case of grasping scissors.
 Fig.~\ref{fig:restoration_from_unknown_interaction_images}(a) shows
 that the shift invariant auto-encoder extracted mask shapes
 regardless of their position.
 Fig.~\ref{fig:restoration_from_unknown_interaction_images}(b) shows
 that the ordinary auto-encoder failed to restore masks accurately;
 however,
 the shift invariant auto-encoder restored typical interaction images.
 The shift invariant auto-encoder can predict a possible typical
 interaction image even from images not used in the training process.

 \subsection{Experiments on Human Hand Imitation using a Robot Hand}
 Here,
 we show examples using the proposed method for controlling a robot
 hand.
 Even though a joint angle can be represented numerically,
 it is not obvious which combinations of joint angles of a robot hand
 are effective for grasping an object.
 If a robot hand can imitate the effective hand posture of a human,
 the cost of designing the joint angles can be reduced.
 Such imitations can be realized by learning the relationship between
 the appearance of a human hand and the corresponding joint angles of a
 robot
 hand.
 A shift invariant auto-encoder can effectively numerically represent
 the appearance of a human hand
 because the corresponding joint angles of a robot hand are independent
 of
 the location of the human hand in the view.

 We developed an imitation program as follows.
 \begin{enumerate}
  \item Collect depth images of postures of a human hand.
  \item Generate a shift invariant auto-encoder for the collected depth
        images.
  \item Allocate the joint angles of a robot hand to several postures of
        a
        human hand.
  \item Train a neural network regressor to calculate the joint
        angles of a robot hand from a descriptor of a human hand
        posture.
 \end{enumerate}
 Using a shift invariant auto-encoder, we can generate a descriptor
 representing a human hand shape directly from the appearances
 without any normalization.

 In this experiment, we used a 3-finger robot hand with 10 degrees
 of freedom (9 DOF fingers and 1 DOF wrist).
 We trained a shift invariant auto-encoder and a regressor so that
 the fingers of the robot hand imitated the thumb, the forefinger and
 the middle finger
 (Fig.~\ref{fig:correspondence_of_fingers}).
 \begin{figure*}[t]
  {\centering
  \begin{minipage}[c]{.68\textwidth}
   {\centering
   \includegraphics[width=.99\textwidth]{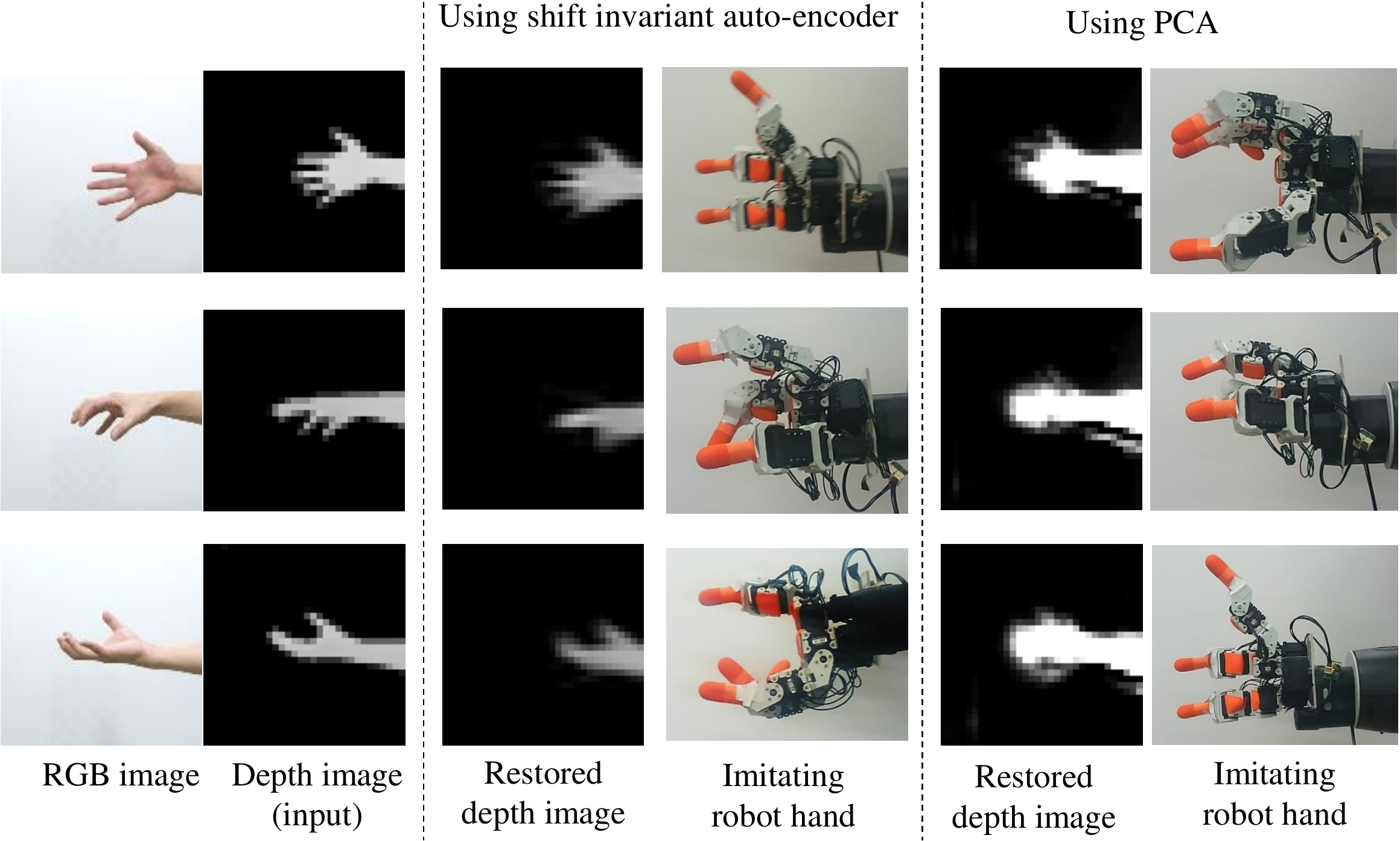}
   \caption{Imitation of a human hand by a robot hand}
   \label{fig:imitation_by_robot_hand}
   }
  \end{minipage}
  \begin{minipage}[c]{.21\textwidth}
   {\centering
  \begin{tabular}{c}
   \includegraphics[width=.89\textwidth]{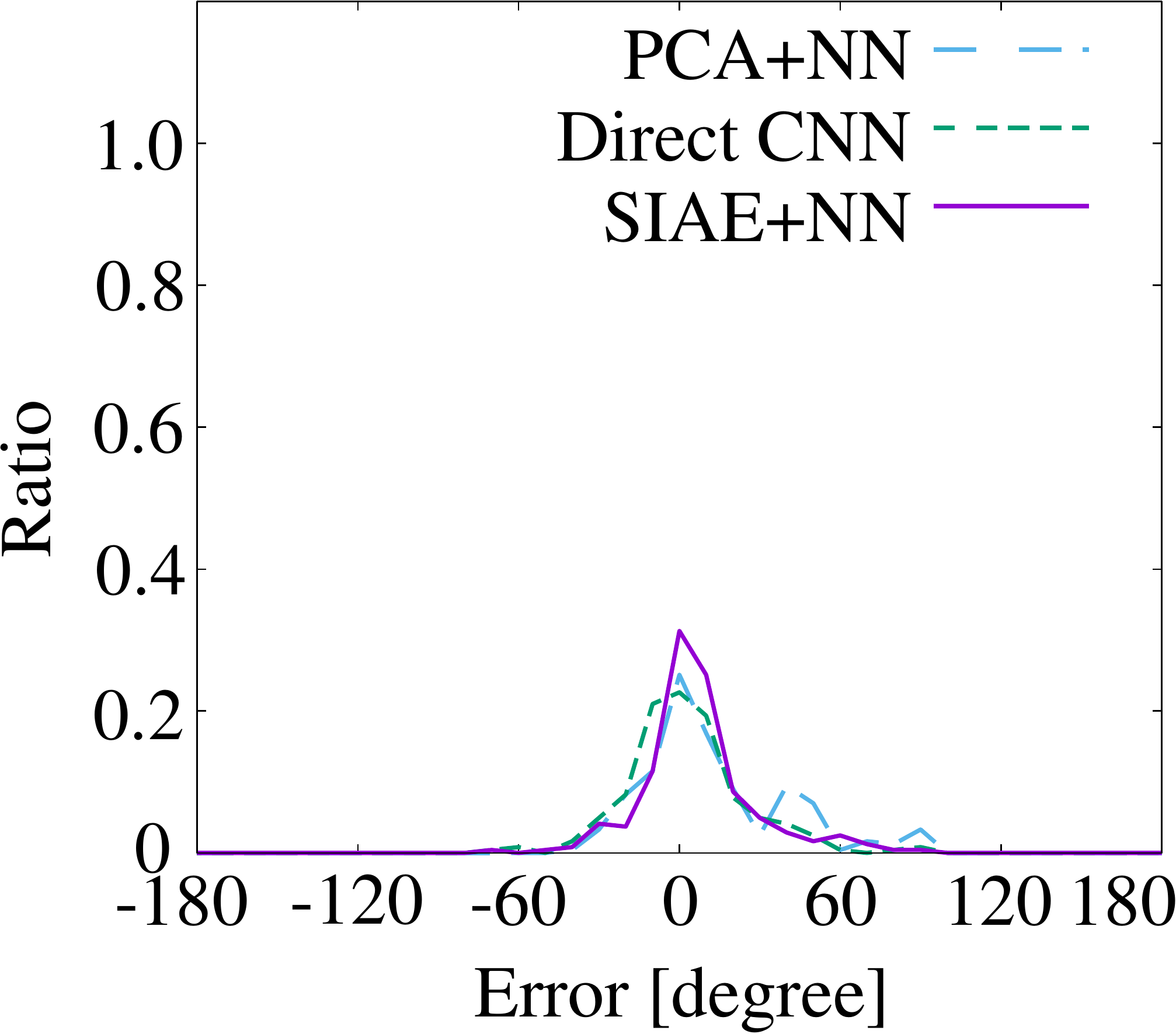}
   \\
   {\footnotesize (a) 9 joints on fingers}
   \\
   \includegraphics[width=.89\textwidth]{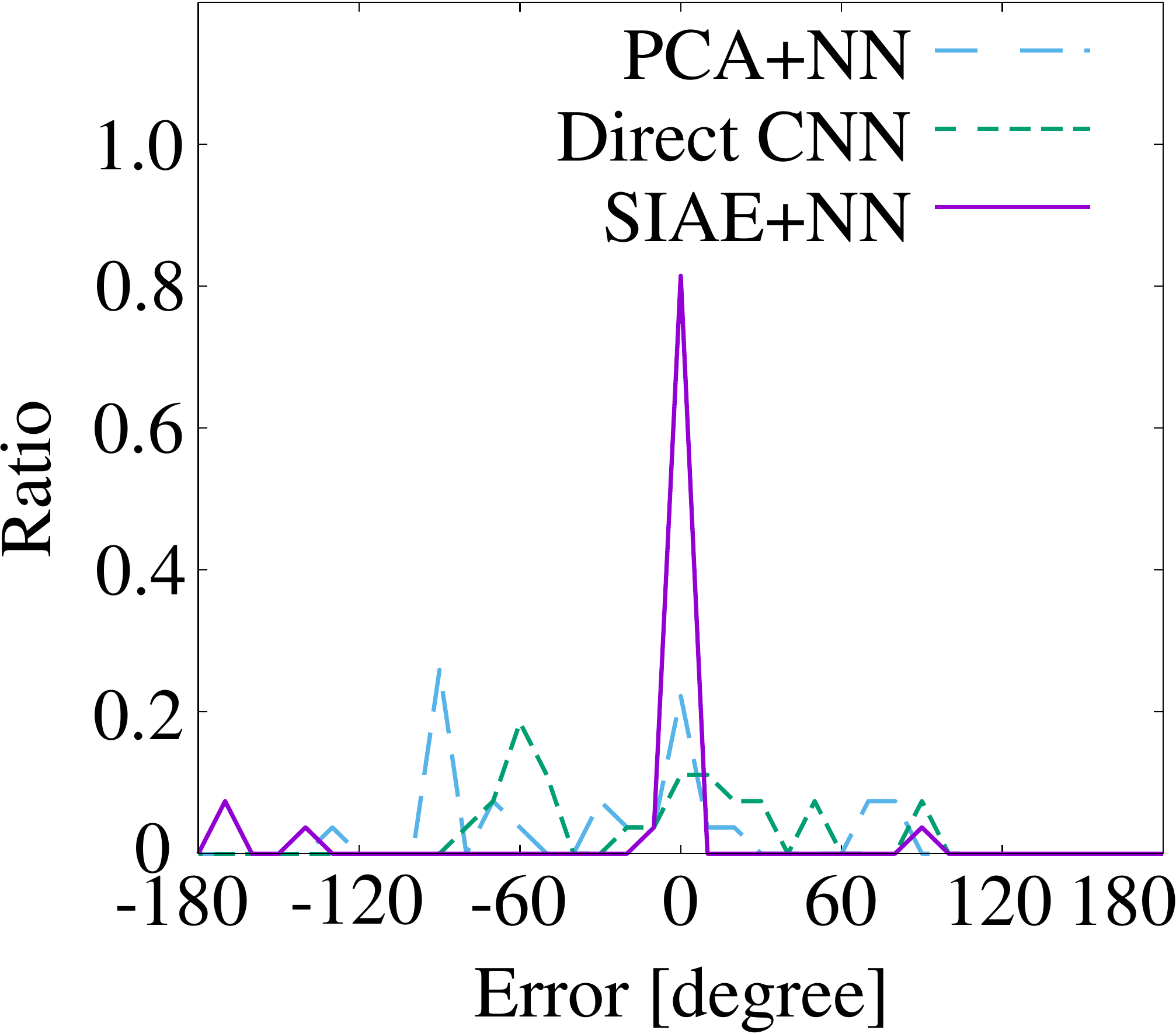}
   \\
   {\footnotesize (b) A joint on the wrist}
  \end{tabular}
  \caption{Error histograms of estimated joint angles}
  \label{fig:histogram_of_estimated_joint_angles}
   }
  \end{minipage}
  }
 \end{figure*}
 The encoder of the shift invariant auto-encoder consists
 of two CNN layers and a three-layer fully connected NN, and the decoder
 is a three-layer fully connected NN.
 In this experiment, the encoder converts a single channel
 $(32\times 32) [\text{pixel}]$ depth image into
 a 100-dimensional descriptor and the regressor converts a descriptor
 into the joint angles of the robot hand.
 The shift invariant auto-encoder is trained with 9377 depth images and
 the regressor is trained with 1339 pairs of descriptors and
 manually determined combinations of joint angles.

 To demonstrate the effectiveness of the proposed method, we
 developed a similar program using principal component analysis
 (PCA) instead of the shift invariant auto-encoder.
 The dimension of the descriptors used by the PCA is 100, which is the
 same as
 that used by the shift invariant auto-encoder.
 The cumulative contribution ratio for the dimension is approximately
 95\%.

 The left two columns in Fig.~\ref{fig:imitation_by_robot_hand}
 show RGB images and depth images of a human hand
 that are not used in the training process.
 The third and fifth columns in the figure show images restored
 by the shift invariant auto-encoder and PCA, respectively.
 The shapes of the fingers are preserved in the images restored by the shift
 invariant auto-encoder; however, PCA does not preserve such features.
 This is because a descriptor generated by PCA depends on the position
 of the
 spatial subpattern and the positions of the human hand in
 Fig.~\ref{fig:imitation_by_robot_hand} are different from the
 training samples.
 This means that the shift invariant auto-encoder is more suitable for
 generating a descriptor representing a spatial subpattern.

 We show the results of the imitations in the fourth and sixth columns in
 Fig.~\ref{fig:imitation_by_robot_hand},
 where the robot hand replays the joint angles calculated by the
 regressor.
 In the case using PCA
 (the sixth column in Fig.~\ref{fig:imitation_by_robot_hand}),
 the joint angles of the robot hand are very different from those of the
 human hand.
 However, in the case using the shift invariant auto-encoder
 (the fourth column in Fig.~\ref{fig:imitation_by_robot_hand}),
 the joint angles of the three fingers of the robot hand appear similar
 to those of the human hand.
 Because a descriptor generated by the shift invariant auto-encoder
 represents the
 spatial shape accurately,
 the regressor can learn the relationship between a shape and the joint
 angles
 more accurately than when using PCA.

 In Fig.~\ref{fig:histogram_of_estimated_joint_angles}, we show
 error histograms for 9 joints on fingers and a joint on the wrist.
 They are calculated from 27 samples that are not used in the training
 process.
 In Fig.~\ref{fig:histogram_of_estimated_joint_angles},
 ``PCA+NN'' and ``SIAE+NN'' mean NN-based angle regressions from a
 descriptor by PCA or Shift Invariant Auto-Encoder (SIAE).
 ``Direct CNN'' means a CNN-based direct regression from
 an input image to joint angles without intermediate descriptors.
 In Fig.~\ref{fig:histogram_of_estimated_joint_angles}(a), the
 histograms are similar, but
 Fig.~\ref{fig:histogram_of_estimated_joint_angles}(b) shows that
 the SIAE+NN estimated the wrist angles more accurately
 than the PCA+NN and the Direct CNN.
 The PCA+NN and the Direct CNN may be overfitted to the training samples.
 The root mean squared errors (RMSEs) of the PCA+NN, the Direct CNN and
 the SIAE+NN for fingers are $30.8, 22.2$ and $22.1 \text{[degree]}$,
 respectively.
 The RMSEs of them for the wrist are
 $64.2, 49.0$ and $56.2\text{[degree]}$, respectively.
 Although the SIAE+NN has a larger RMSE due to a few samples with very
 large error, the ratio of samples with errors within
 $\pm 5\text{[degree]}$ is 81\% much larger than
 22\% (the PCA) and 11\% (the Direct CNN)
 as shown in Fig.~\ref{fig:histogram_of_estimated_joint_angles}(b).

 These results show that a shift invariant auto-encoder is effective for
 regression based on a spatial subpattern.

 \section{CONCLUSIONS}
 We proposed a transform invariant auto-encoder and demonstrated that a
 shift invariant auto-encoder  can generate a
 descriptor representing a spatial subpattern regardless of
 its position.
 In several experiments, we showed that the proposed method is
 applicable
 to regression based on a spatial subpattern.

 In this paper, we experimented with spatial subpatterns and shifts.
 However, the framework of the proposed cost function can be applied to
 temporal patterns and other transforms such as dilation and
 rotation.
 Since the proposed function requires enumeration of transforms, random
 sampling of transforms may be required to suppress the computation
 cost.
 With such an extension, an auto-encoder will be able to independently
 encode typical motions in a video without regard to dilation and
 rotation.
 This will be useful for motion-based recognition.

%

\addtolength{\textheight}{-12cm}   



%

%



\bibliographystyle{IEEEtran}
\bibliography{mybibliography}
\end{document}